\documentclass{article}

\PassOptionsToPackage{numbers, compress}{natbib}

 \usepackage[preprint]{neurips_2026}

\usepackage[utf8]{inputenc} %
\usepackage[T1]{fontenc}    %
\usepackage{url}            %
\usepackage{booktabs}       %
\usepackage{amsfonts}       %
\usepackage{nicefrac}       %
\usepackage{microtype}      %
\usepackage[table]{xcolor}  %
\usepackage{amsmath}
\usepackage{amssymb}
\usepackage{graphicx}
\usepackage{wrapfig}
\usepackage{algorithm}
\usepackage{algpseudocode}
\usepackage{float}
\usepackage{placeins}
\usepackage{xfrac}
\usepackage{hyperref}       

\definecolor{lightgreen}{RGB}{200,225,200}
\definecolor{lightred}{RGB}{252,230,225}
\definecolor{lightblue}{RGB}{225,235,246}
\definecolor{lbsfg}{RGB}{162,52,58}
\definecolor{lpsfg}{RGB}{42,78,142}
\definecolor{edgreen}{RGB}{0,110,45}

\hypersetup{
    colorlinks=true,
    filecolor=magenta,
    urlcolor=black,
    citecolor=brown,
}

\usepackage{enumitem}
\setlist[itemize]{leftmargin=10mm}

\newcommand{\new}[1]{\textcolor{black}{#1}}
\newcommand{\mg}[1]{\textcolor{black}{#1}}
\newcommand{\pp}[1]{\textcolor{black}{#1}}
\newcommand{\ed}[1]{\textcolor{black}{#1}}
\newenvironment{revblock}{\color{black}\ignorespaces}{\ignorespacesafterend}
\providecommand{\argmax}{\operatorname*{arg\,max}}

\newcommand{\tabincell}[2]{\begin{tabular}{@{}#1@{}}#2\end{tabular}} 
\newcommand{\email}[1]{\href{mailto:#1}{\texttt{#1}}}

\title{Early-Bird Decoding: Accelerating Diffusion LLMs with Learnable Block Sizes and Parallel Sampling}

\author{%
  Lixuan Wei$^{1}$\thanks{Equal contribution. Work done while interning at Purdue EcoAI Lab.},\,
  Wei Zhou$^{2}$\footnotemark[1],\,
  Jianwen Wu$^{3}$,\,
  Yipeng Shen$^{3}$,\,
  Meiling Wang$^{3}$,\,
  Haoran You$^{3}$\thanks{Corresponding author. Email: \email{haoran@purdue.edu}} \\[0.6em]
  $^{1}$Harvard University \quad
  $^{2}$Georgia Institute of Technology \quad
  $^{3}$Purdue University \\
}

\begin{document}

\maketitle

\begin{abstract}

Diffusion large language models (dLLMs) offer a promising parallel decoding paradigm as an alternative to autoregressive generation through iterative unmasking.
However, dLLMs typically require many steps before token confidence reaches the decoding threshold, resulting in inefficient inference even with block-wise KV caching.
To accelerate dLLM inference, we \textit{for the first time} propose an ``early-bird (EB)'' decoding framework, motivated by the observation that tokens with similarly low entropy tend to cluster and can be jointly decoded earlier, before reaching the confidence threshold.
In particular, our EB-Decode framework integrates two key enablers:
(1) a learnable network that adaptively groups tokens with similar uncertainty into variable-length blocks, rather than relying on fixed block sizes;
(2) a position-aware sampler that learns to unmask tokens in parallel using fewer decoding steps within predicted variable-length blocks.
Both components are developed without modifying pretrained dLLM weights and can therefore be directly deployed as plug-ins during serving, with negligible training and inference overhead.
Extensive experiments across three models and four benchmarks consistently validate our observation and the effectiveness of EB-Decode, achieving \textbf{\ed{3.53--18.76$\times$}} higher throughput than the vanilla decoding method and up to \textbf{1.58$\times$} higher throughput over the strongest baseline, Fast-dLLM, with comparable accuracy.
\end{abstract}

\section{Introduction}
\label{sec:introduction}
Diffusion large language models (dLLMs) have emerged as a compelling alternative to autoregressive (AR) models, breaking the sequential bottleneck of token-by-token decoding by generating text in parallel through iterative denoising over masked positions~\citep{austin2021structured,sahoo2024simple,nie2025large,ye2025dream}.
This paradigm enables bidirectional context utilization and makes dLLMs a practical and increasingly scalable approach for fast, high-quality text generation, as demonstrated by large-scale systems such as Gemini Diffusion~\citep{gemini_diffusion}, Seed-Diffusion~\citep{song2025seed}, and Mercury~\citep{khanna2025mercury}.
However, current open-source dLLMs such as LLaDA~\citep{nie2025large,zhu2025lladamoe,bie2025llada2} and Dream~\citep{ye2025dream} still struggle to consistently deliver better accuracy-efficiency tradeoffs compared to their AR counterparts of similar size; for example, 
the AR model \ed{LLaMA3-8B-Instruct} achieves 48.0 tokens/s, whereas the dLLM \ed{LLaDA-8B-Instruct} reaches only 3.5 tokens/s under an NVIDIA A100-PCIe
40GB GPU setup. Its accuracy also drops from 83.5\% at 1,024 denoising steps to 54.1\% at 256 denoising steps~\citep{wang2025diffusion,peng2025efficient,kang2025parallelbench,li2026diffusion}, which limits the widespread adoption of dLLMs in real-world deployments.

Recently, several works have attempted to mitigate the inference efficiency challenges of dLLMs, including semi-AR variants for block-wise sequential generation~\citep{arriola2025block,cheng2025sdar}, Key-Value (KV) and activation caching~\citep{wu2025fast,liu2025dllm,ma2025dkv,hu2025flashdlm,chen2025dpad}, efficient sampling algorithms~\citep{huang2025pcsampler,bao2025learning,wei2025accelerating,lu2025adablock,israel2025accelerating,wang2025time,zhang2026swordsman,luo2026dsb}, and step distillation~\citep{hayakawa2024Di4C,deschenaux2024SDTT,chen2025dparallel,liang2026cd4lm,zhang2026t3d}. However, most of them rely on fixed block selection and confidence thresholds during decoding, implicitly assuming uniform token difficulty along the sequence, which wastes parallelism on hard tokens and under-utilizes it on easy ones, contradicting the heterogeneous nature of natural language. For example, in code, function signatures are typically easier to predict than function bodies, and in mathematical reasoning, routine arithmetic is straightforward whereas multi-step derivations are more challenging. 
Beyond that, a few recent methods introduce simple heuristics for adaptive block sizing and thresholding, such as delimiter (e.g., period) detection~\citep{lu2025adablock}, sliding windows~\citep{luo2026dsb}, or entropy-based boundaries~\citep{zhang2026swordsman}. Some methods attempt to learn parallel decoding strategies~\citep{bao2025learning}, but still rely on fixed block sizes. These limitations call for a principled framework that not only adaptively determines block sizes but also inherently enables early decoding under homogeneous token difficulty within each block.

\begin{figure*}[t]
    \centering
    \includegraphics[width=1\linewidth]{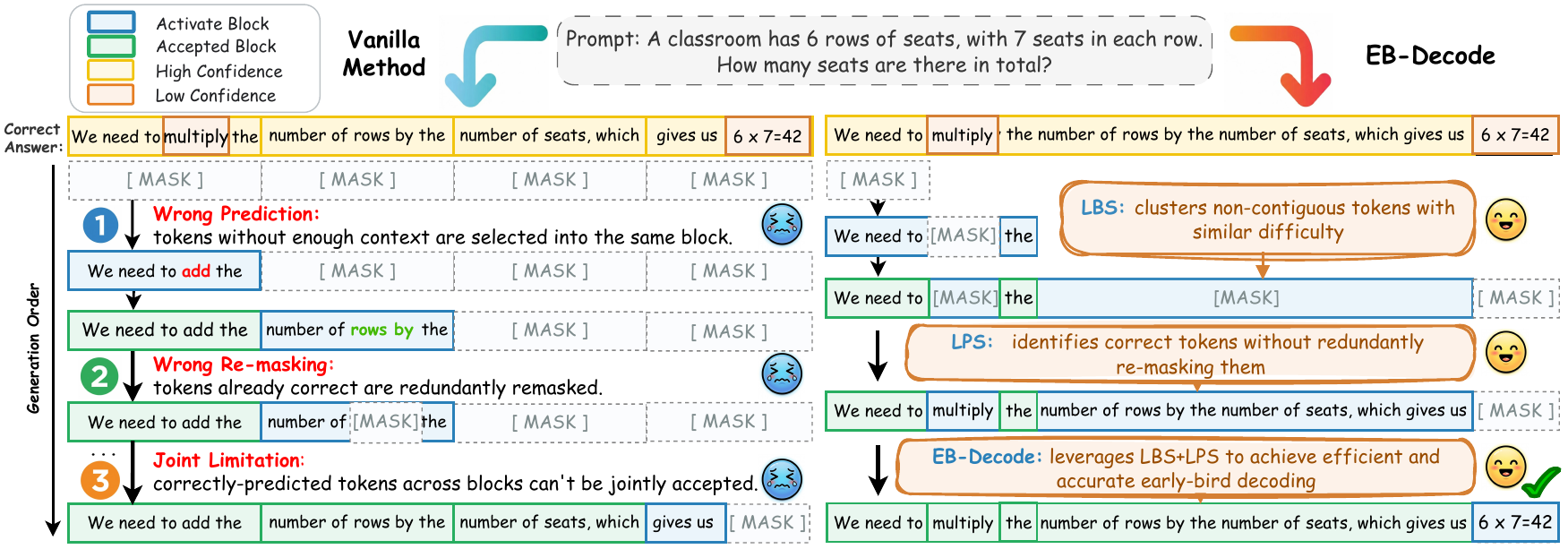}
    \caption{Conceptual comparison illustrating the differences between vanilla block-wise decoding (Left)~\citep{arriola2025block} and the proposed EB-Decode framework (Right).}
    \label{fig:teaser}
\end{figure*}

In this work, we \textit{for the first time} propose a principled ``early-bird'' decoding framework, motivated by our observation that tokens with similar low entropy and close semantic meanings tend to cluster and can be jointly decoded much earlier before reaching the confidence threshold, due to their similar difficulty levels. To enable such principled EB decoding, two key challenges arise:
\textbf{First}, how can we automatically learn adaptive block sizes instead of relying on heuristics? Unlike prior delimiter-based methods, our observation shows that token entropy exhibits a staircase pattern across token positions, indicating that at certain key steps, adjacent or even non-contiguous tokens share similar entropy or uncertainty. This motivates the design of a lightweight network that leverages entropy and positional semantics to dynamically predict block sizes on the fly.
\textbf{Second}, how can we decode the predicted block of tokens earlier before reaching the confidence threshold? Our observation shows that confidence-based decoding wastes computation on tokens that have already converged: many tokens become correct in early denoising steps but are repeatedly remasked because their confidence has not yet reached the threshold. Moreover, under fixed block sizes, confident tokens outside the current block are excluded from early finalization. These observations motivate a position-aware, learnable parallel sampler that leverages per-token statistics (i.e., entropy, position, and step) to determine which positions can be finalized at earlier steps, while naturally supporting variable-length blocks.
To the best of our knowledge, this work is the first to tackle the above challenges toward a principled EB decoding framework. Our contributions are summarized as follows:

\begin{itemize}
    \item We propose a principled early-bird decoding framework for efficient dLLM inference acceleration, termed \textbf{EB-Decode}, that automatically clusters tokens with similar difficulty into blocks and enables their early decoding with fewer denoising steps than confidence-based methods.
    \item \textit{Enabler 1}: We adopt a \textbf{learnable block size (LBS)} prediction network to dynamically cluster contiguous or non-contiguous tokens with similar difficulty and semantic coherence on the fly.
    \item \textit{Enabler 2}: We introduce a position-aware \textbf{learnable parallel sampling (LPS)} network to identify and finalize converged tokens at early denoising steps under variable-length blocks.
    \item Extensive experiments across three dLLMs and four representative benchmarks demonstrate the effectiveness of EB-Decode: it achieves \textbf{\ed{3.53--18.76$\times$}} throughput improvement over the vanilla decoder with comparable accuracy and only \textbf{$\sim$6.21\%} routing overhead, and \ed{delivers} an additional \textbf{1.20$\times$} speedup when combined with KV-cache pipelines.
\end{itemize}

\section{Related Works}
\label{sec:related}

\textbf{dLLMs.} Unlike AR models, dLLMs generate text through an iterative denoising process over discrete tokens~\citep{austin2021structured,sahoo2024simple,shi2024simplified,lou2023discrete,zheng2025masked}, achieving likelihood comparable to their AR counterparts.
At the billion-parameter scale, LLaDA~\citep{nie2025large} performs on par with LLaMA3, and subsequent extensions further improve its alignment, sparsity, and scaling~\citep{zhu2025llada,zhu2025lladamoe,bie2025llada2}. Dream-7B~\citep{ye2025dream} adds AR-based initialization and adaptive noise rescheduling.
We provide more literature review of dLLM in Appendix~\ref{app:dllm_related}.

\textbf{Efficient Inference of dLLMs.}
Recent work improves dLLM decoding along several directions. 
Cache-based methods reuse key-value projections across denoising steps to avoid redundant computation~\citep{liu2025dllm,wu2025fast,ma2025dkv,hu2025flashdlm}. 
Distillation-based methods train the model to commit more tokens per step~\citep{chen2025dparallel,liang2026cd4lm}. 
Efficient sampling methods redesign the denoising trajectory to better allocate computation across steps.
For example, 
SlowFast~\citep{wei2025accelerating} alternates between exploratory and accelerated phases; 
DUS~\citep{luxembourg2025plan} front-loads computation to early steps via dilated scheduling;
Adaptive acceptance methods replace the static confidence threshold with learned or per-position confidence~\citep{israel2025accelerating,bao2025learning,ma2025dinfer,li2025diffusion}. 
Dynamic block methods resize the block at runtime by aligning boundaries with confidence or entropy shifts~\citep{lu2025adablock,zhang2026swordsman,luo2026dsb}. 
An orthogonal line targets test-time quality through revocable draft-and-verify decoding~\citep{hong2025wide}, temporal-dynamics voting across denoising steps~\citep{wang2025time}, and inference-time remasking~\citep{wang2025remasking}. 
In contrast, our proposed principled EB-Decode explicitly leverages the token-level statistics, and learns to dynamically cluster tokens with similar difficulty and adapt the sampling schedule to identify and finalize converged tokens at early denoising steps under variable-length blocks.

\textbf{Early-Bird (EB) Phenomenon.}
Early prediction is important for efficient training and inference. 
For training, the EB ticket hypothesis~\citep{you2019drawing} shows that small subnetworks (i.e., lottery tickets~\citep{frankle2018lottery}) can be identified early in training via mask-distance convergence, achieving accuracy comparable to overparameterized networks. 
This EB phenomenon has been consistently observed in BERT~\citep{chen2021earlybert}, GCNs~\citep{you2022early}, LLMs~\citep{gu-etal-2024-light}, and diffusion models~\citep{whalen2025early}.
For inference, early-exit mechanisms enable dynamic computation by terminating inference once predictions become sufficiently confident. 
They were first applied to RNNs through a learned halting unit~\citep{graves2016adaptive}, and later generalized to CNNs~\citep{teerapittayanon2016branchynet} and Transformers~\citep{xin2020deebert,zhou2020bert,schuster2022confident}, where intermediate layers can halt computation once a confidence or prediction-agreement criterion is satisfied.
In this work, we \textit{for the first time} observe the EB phenomenon during the decoding phase in dLLMs and leverage it to enable efficient EB decoding.

\section{Preliminaries of dLLMs}
\label{sec:preliminaries of dLLMs}
dLLMs formulate text generation as a diffusion process over discrete token sequences, consisting of a forward masking process and a reverse denoising process. Let $\mathbf{x}_0 = (x^1_0, \ldots, x^L_0)$ denote a clean token sequence of length $L$ drawn from a vocabulary $\mathcal{V}$, where $x^i_0$ is the token at position $i$. The forward process samples a noise level $t \in [0,1]$ and produces a corrupted sequence $\mathbf{x}_t = (x^1_t, \ldots, x^L_t)$ by independently replacing each token with the special \texttt{[MASK]} symbol with probability $t$, so that $\mathbf{x}_t$ converges to a fully masked sequence as $t \rightarrow 1$. The reverse process is parameterized by a bidirectional Transformer that defines a mask predictor $p_\theta(\cdot \mid \mathbf{x}_t)$, which recovers the distribution over the original token at every masked position simultaneously. The model is trained by minimizing
\begin{equation}
    \mathcal{L}(\theta) = \mathbb{E}_{t \sim \mathcal{U}(0,1),\, \mathbf{x}_0,\, \mathbf{x}_t} \left[ \frac{1}{t} \sum_{i=1}^{L} \mathbf{1}_{\{x^i_t = \texttt{[MASK]}\}} \bigl( - \log p_\theta(x^i_0 \mid \mathbf{x}_t) \bigr) \right],
\end{equation}
where $\mathbf{1}_{\{\cdot\}}$ is the indicator function that restricts the cross-entropy to currently masked positions, and the factor $1/t$ compensates for the expected mask ratio at noise level $t$. This objective upper-bounds the negative log-likelihood of the model distribution.

\textbf{Block-wise Decoding.}
At inference time, dLLMs commonly adopt a block-wise decoding strategy~\citep{arriola2025block,wu2025fast}: the response is partitioned into contiguous fixed-size blocks decoded from left to right, where positions inside the active block are predicted in parallel and committed based on confidence, while the rest are remasked for further refinement. The full update rule is deferred to Appendix~\ref{app:dllm_prelim}.
However, reliance on fixed block sizes and static commit rules ignores variation in token difficulty along the sequence, motivating our empirical analysis in Sec.~\ref{sec:observation of EB Phenomenon in dLLM Decoding}.

\section{The Observations of EB Phenomenon in dLLM Decoding}
\label{sec:observation of EB Phenomenon in dLLM Decoding}

To understand the inefficiency of existing dLLM decoding, we analyze the denoising trajectory and identify two empirical phenomena overlooked by fixed-block, threshold-based decoding: a \textit{staircase entropy pattern} across token positions and an \textit{early convergence phenomenon} across denoising steps.

\textbf{Observation 1: Staircase Entropy Pattern.}
To examine how token difficulty varies across positions, we measure the Shannon entropy $H^i_t$ at each masked position $i$ and denoising step $t$. 
We visualize this on a GSM8K sample with generation length 256, 
decoded by the block-wise diffusion method~\citep{wu2025fast}. 
\begin{wrapfigure}{r}{0.5\textwidth}
    \centering
    \includegraphics[width=\linewidth]{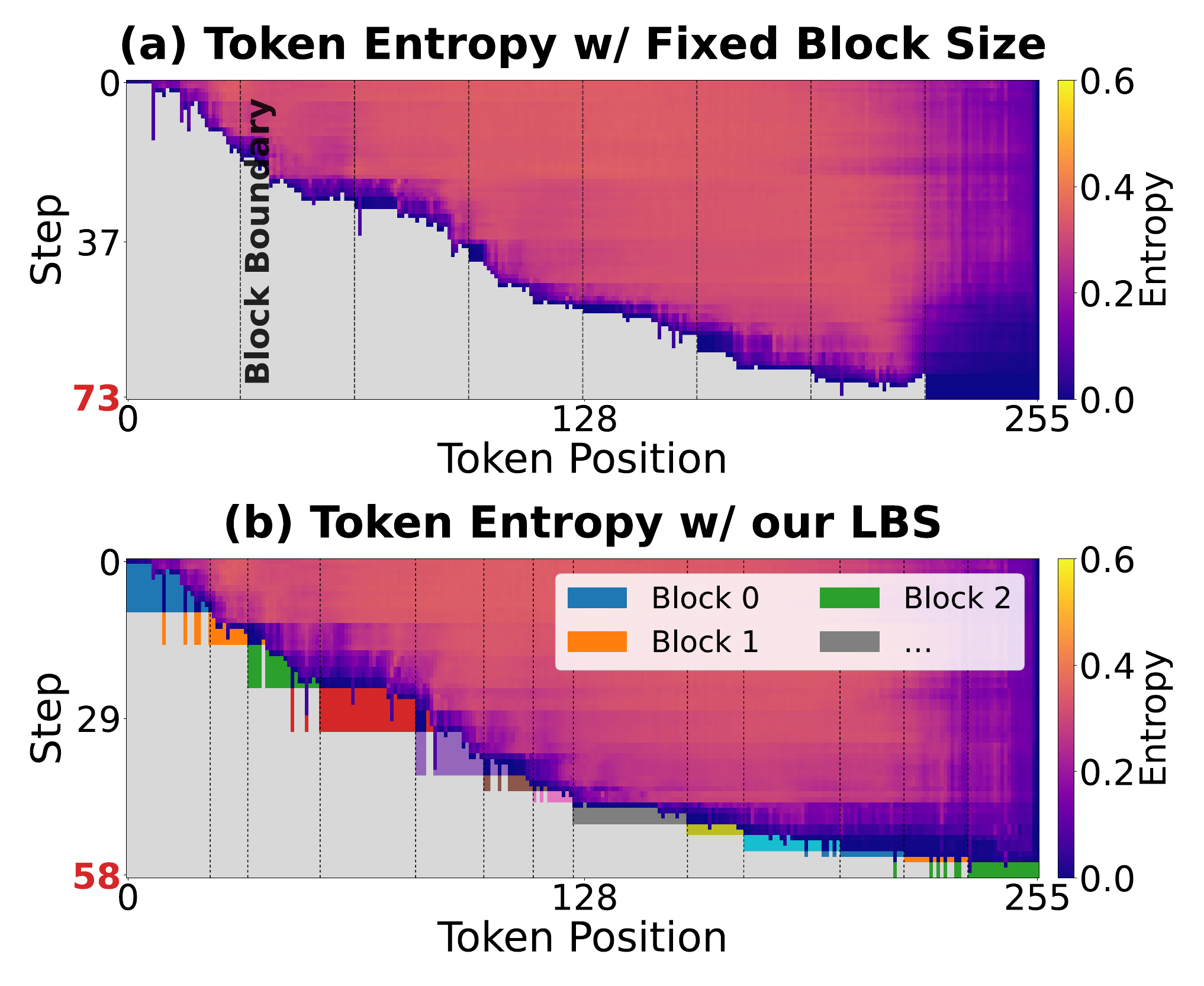}
    \caption{Comparison of entropy heatmaps between \textbf{(a)} fixed block size and \textbf{(b)} our non-contiguous learnable block size method, where each colored region represents a block. Gray regions denote tokens that have been unmasked.}
    \label{fig:entropy}
\end{wrapfigure}
As shown in Fig.~\ref{fig:entropy} (a),
the entropy does not decrease smoothly across token positions. Instead, it exhibits a staircase pattern, consisting of long, flat low-entropy plateaus that typically correspond to predictable syntax or boilerplate content, such as reasoning connectives like \textit{``To solve this problem, we first need to \ldots''} in mathematical reasoning tasks, or structural scaffolding around known function signatures like \texttt{def function\_name(args):} in code generation tasks. This suggests that semantically coherent tokens within each plateau share similar difficulty, and these regions exhibit similarly low entropy across denoising steps. However, this conflicts with fixed block boundaries, under which such semantically coherent regions are split in a manner agnostic to semantic structure, preventing stabilized regions across block boundaries from being decoded jointly and leaving the available parallelism unexploited.
In addition, these flat plateaus are occasionally interrupted by short high-entropy spikes corresponding to harder-to-predict tokens. Although such hard tokens sit within an otherwise semantically coherent region, they are not well-suited to being decoded jointly with the surrounding easy tokens. Yet fixed-block methods enforce a strict left-to-right order that forces these hard tokens to be resolved inside the current block before any subsequent content can be generated. In contrast, as shown in Fig.~\ref{fig:entropy} (b), a block selection strategy that allows non-contiguous grouping could defer these hard tokens to a later block, letting the model first decode the surrounding easy context and leverage the additional right-side information when eventually resolving them.

\begin{figure}[h]
    \centering
    \begin{minipage}[c]{0.51\textwidth}
        \centering
        {\small \textbf{(a) Per-token Denoising Trajectory}}\par
        \vspace{0.2em}
        \includegraphics[width=\linewidth]{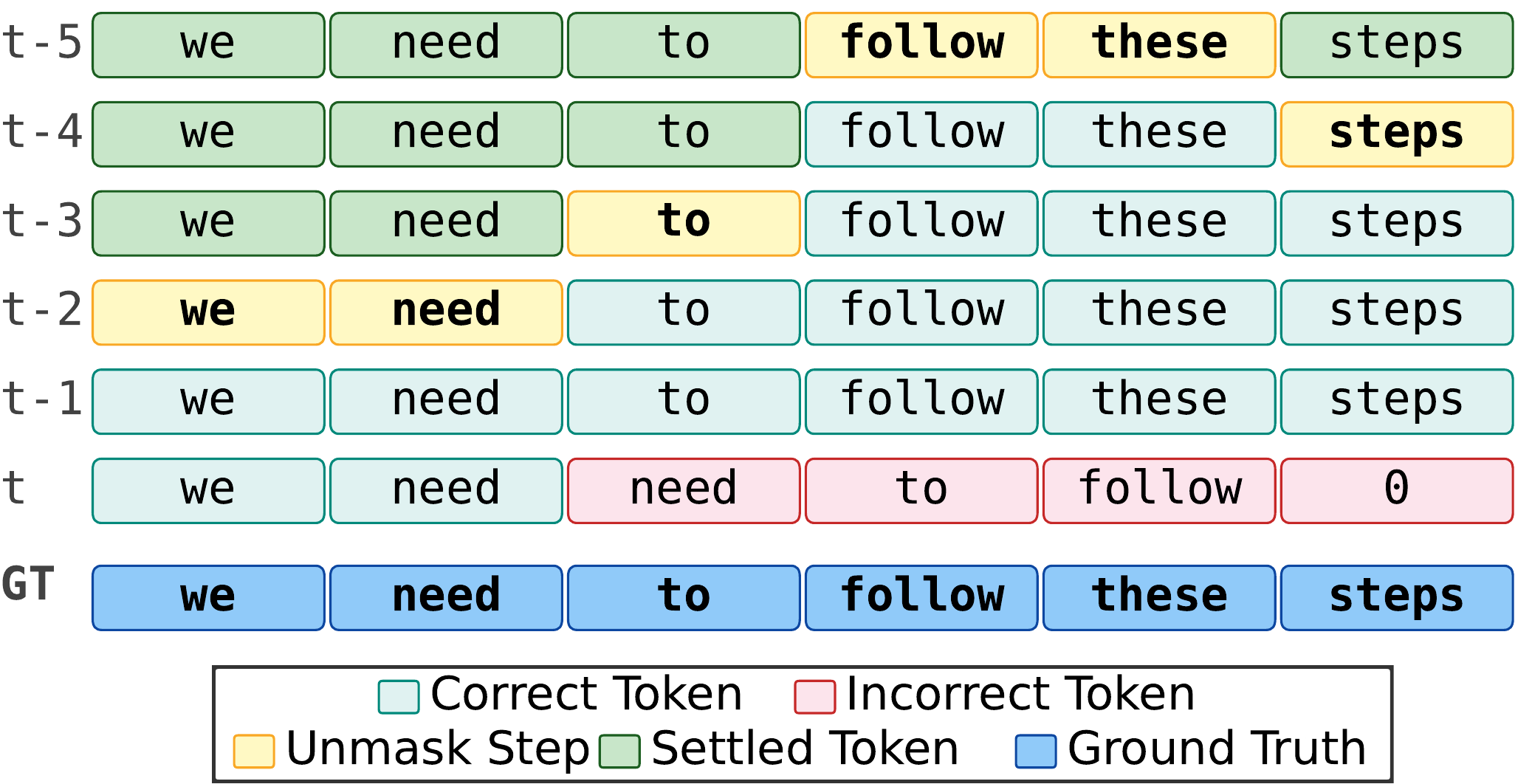}
    \end{minipage}\hfill
    \begin{minipage}[c]{0.48\textwidth}
        \centering
        {\small \textbf{(b) Block-relative Lead-time on GSM8K}}\par
        \vspace{0.4em}
        \footnotesize
        \setlength{\tabcolsep}{4pt}
        \begin{tabular}{l|c|c|c}
        \toprule
        & Vanilla~\citep{nie2025large} & \tabincell{c}{Confidence-\\based~\citep{wu2025fast}} & \tabincell{c}{\textbf{EB-Decode}\\\textbf{(Ours)}} \\
        \midrule
        Mean Time       & 13.4 Steps & 2.3 Steps & \textbf{0.7 Steps} \\
        \midrule
        \multicolumn{4}{c}{\textit{Fraction of Tokens}} \\
        \midrule
        $\geq$1 Step     & 96.3\%  & 62.5\% & \textbf{25.1\%} \\
        $\geq$3 Steps     & 88.3\%  & 35.0\% & \textbf{10.5\%} \\
        $\geq$5 Steps    & 80.2\%  & 18.1\% & \textbf{4.6\%}  \\
        $\geq$10 Steps   & 60.5\%  & 3.5\%  & \textbf{0.7\%}  \\
        \bottomrule
        \end{tabular}
    \end{minipage}
    \caption{
    \textbf{(a)} Per-token denoising trajectory, where each column represents a token position across steps within one segment. 
    Light green cells indicate tokens whose predictions become correct before they are committed, illustrating the inefficiency of confidence-based decoding.
    \textbf{(b)} Mean lead-time (in denoising steps) and fraction (\%) of tokens with lead $\geq k$ denoising steps before commit.
    }
    \label{fig:early_decode}
\end{figure}

\textbf{Observation 2: Early Convergence Phenomenon.}
Even when block selections are well aligned with token difficulty, decoding still has to decide \emph{when} each token within a block should be committed. To examine this, we track the per-step prediction at each masked position for six representative token positions of a GSM8K sample decoded using the vanilla block-wise method~\citep{nie2025large}. As shown in Fig.~\ref{fig:early_decode} (a), we observe that many tokens reach the correct prediction several denoising steps before their per-token confidence crosses the threshold, in principle allowing them to be committed early. These early-converged tokens, however, are repeatedly remasked before being finalized, introducing step-level redundancy. 
Fig.~\ref{fig:early_decode} (b) quantifies this redundancy by showing that under both the vanilla~\citep{nie2025large} and confidence-based method~\citep{wu2025fast}, many tokens remain unaccepted for multiple denoising steps after their predictions become correct. 
As reported in Fig.~\ref{fig:early_decode} (b), the mean lead times of the two baseline methods are 13.4 and 2.3 steps, respectively. 
The fractions of tokens are 96.3\% and 62.5\% when the lead time is $\geq 1$ step, and 60.5\% and 3.5\% when the lead time is $\geq 10$ steps. 
This behavior arises because existing methods evaluate each masked position independently and cannot distinguish correct-but-uncertain predictions from genuinely incorrect ones, highlighting the need for a principled and learnable strategy to commit correct tokens earlier.

\begin{figure*}[!t]
    \centering
    \includegraphics[width=1\linewidth]{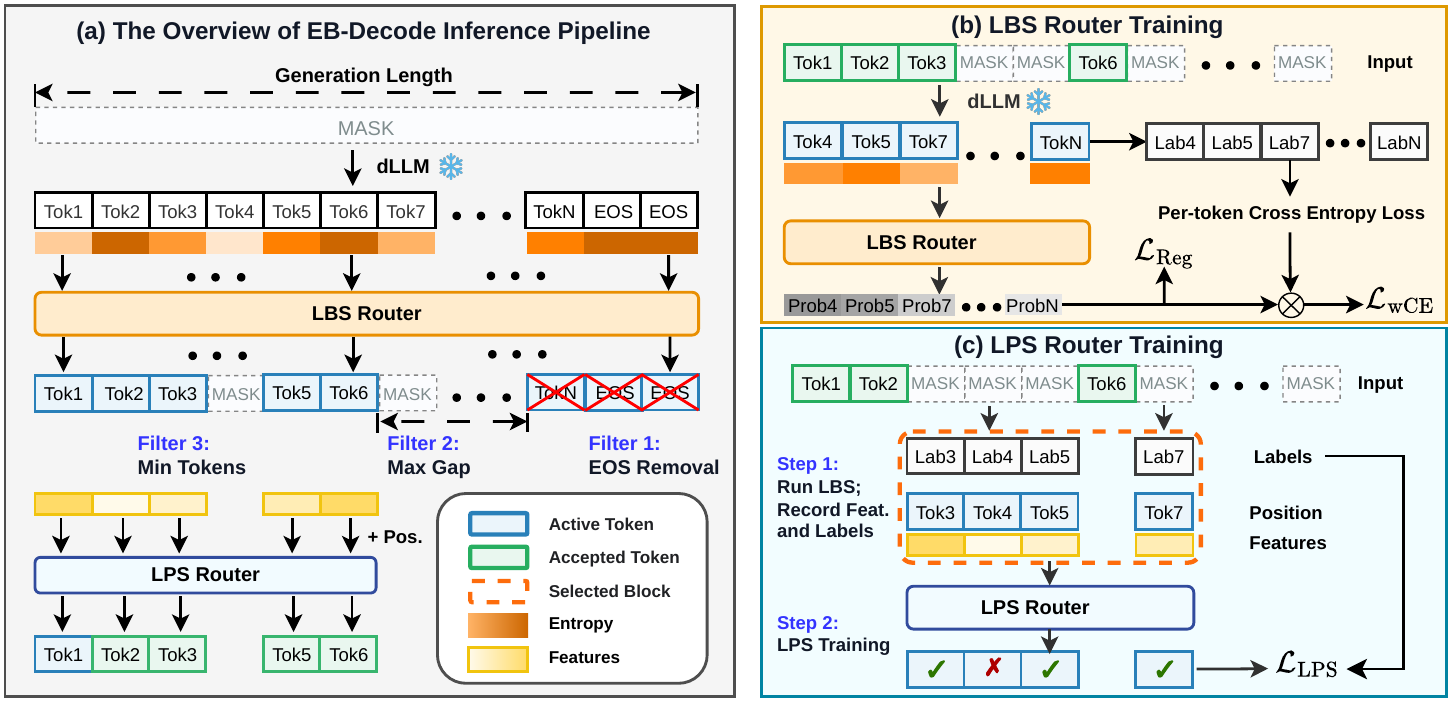}
    \caption{Overview of the proposed EB-Decode framework. (a) Inference pipeline: LBS first predicts and filters the active block out, and LPS then selects the tokens to commit within this block. (b) LBS router training with a weighted cross-entropy loss and a regularization term. (c) LPS router training, where features and labels are first recorded by decoding with LBS and then used to train the router.}
    \label{fig:overview}
\end{figure*}

\section{The Proposed EB-Decode Framework}

\textbf{Overview.}
The two observations in Sec.~\ref{sec:observation of EB Phenomenon in dLLM Decoding} reveal two fundamental issues in current dLLM decoding: 
(1) a fixed block size that ignores the staircase pattern of token difficulty and the non-contiguous nature of tokens that should be grouped, and (2) a fixed confidence threshold that overlooks the adaptive nature of how early predictions actually converge.
As illustrated in Fig.~\ref{fig:overview} (a),
our proposed EB-Decode framework replaces the aforementioned two static choices with learnable counterparts while keeping the base dLLM frozen. 
Specifically, LBS replaces the fixed block size and decides \emph{where} to decode at each step, as described in Sec.~\ref{sec:adablock}, while LPS replaces the fixed confidence threshold and decides \emph{which} positions within a given variable-length block to finalize early, as described in Sec.~\ref{sec:LPS}.
Both modules reuse the base dLLM's forward pass and introduce negligible training and inference overhead.

\subsection{Enabler 1: Learnable Block Size (LBS)}
\label{sec:adablock}

The staircase entropy pattern identified in Sec.~\ref{sec:observation of EB Phenomenon in dLLM Decoding} reveals that token difficulty is spatially clustered: low-entropy regions alternate with sharp uncertainty spikes that do not align with fixed block boundaries.
A fixed partition either splits an easy plateau across two blocks, wasting a decoding pass, or lumps easy and hard tokens together, causing hard tokens to bottleneck their easier neighbors.
To resolve this, we introduce LBS, a lightweight router $\mathcal{R}_\phi$ that at each step selects a variable-length, non-contiguous active block $\mathcal{B} \subseteq \mathcal{M}$ from the currently masked positions $\mathcal{M} = \{i : x^i = \texttt{[MASK]}\}$.

\textbf{LBS Router.}
As shown in Fig.~\ref{fig:overview} (b),
for each masked position $i$, the router takes two complementary inputs:
(1) \emph{The normalized entropy}, defined as $\tilde{H}^i = \tilde{H}(p^i) = (-\sum_{v \in \mathcal{V}} p^i(v)\log p^i(v))/\log |\mathcal{V}|$, which measures local prediction uncertainty.
Entropy alone, however, does not distinguish between qualitatively different sources of uncertainty: a function word may be uncertain among a few interchangeable alternatives, while a rare content word may be uncertain because its identity depends on context not yet resolved\ed{---}two situations that call for different routing decisions;
(2) \emph{The top-1 predicted token IDs ($\hat{x}^i$)}.
Inspired by \ed{\citet{lu2025adablock}}, we embed token IDs using the frozen base-model token embedding table. This allows the router to differentiate between semantically trivial tokens and semantically informative ones.
In summary, the router maps these input signals to a per-masked-token inclusion probability $p^i_\phi$ of whether a masked token should be included in the final selected block:
\begin{equation}
    p^i_\phi = \sigma(O^i_\phi) = \sigma(\mathcal{R}_\phi(\tilde{H}^i, \hat{x}^i)) \in (0,1), i \in \mathcal{M},
\end{equation}
where $\sigma$ is the sigmoid function. The router is implemented as a lightweight two-layer Transformer encoder, with full architectural details described in Appendix~\ref{app:router_arch}.

\textbf{LBS Router Training.}
We train the LBS router in a block-wise masking style. A single frozen forward pass through the dLLM yields entropy $\tilde{H}^i$ and token ID $\hat{x}^i$ at every masked position, which are then passed into the LBS router to get the per-token probability $p^i$.
We train the router by minimizing a weighted cross-entropy loss ($\mathcal{L}_{\mathrm{wCE}}$) with a regularization term ($\mathcal{L}_{\mathrm{Reg}}$):
\begin{equation}
    \mathcal{L}_{\mathrm{LBS}}
    =
    \mathcal{L}_{\mathrm{wCE}} - \mathcal{L}_{\mathrm{Reg}}
    =
    \frac{1}{|\mathcal{M}|}
    \sum_{i \in \mathcal{M}}
    \sigma(O^i_\phi) \cdot \ell^i_{\mathrm{CE}}
    \;-\;
     \frac{1}{|\mathcal{M}|}\,\displaystyle\sum_{i \in \mathcal{M}} \sigma(O^i_\phi)
    \label{eq:lbs_loss}
\end{equation}
where $\ell^i_{\mathrm{CE}}$ denotes the per-masked-token cross-entropy loss of the frozen dLLM at position $i$, and $O^i_\phi=\mathcal{R}_\phi(\tilde{H}^i, \hat{x}^i)$ is the router output.
The first term $\mathcal{L}_{\mathrm{wCE}}$ trains the router to assign high inclusion probability (i.e., likelihood of being included in the current block) to positions where the dLLM already predicts correctly (i.e., low $\ell^i_{\mathrm{CE}}$), and assign low inclusion probability to positions where the dLLM is either incorrect or underconfident (i.e., high $\ell^i_{\mathrm{CE}}$).
The second term acts as a regularization role that prevents the degenerate solution of selecting no tokens\ed{---}which would trivially minimize the first term\ed{---}by rewarding larger total inclusion; the subtraction ensures the regularizer and the loss pull in opposite directions, stabilizing the block size.
The training procedure is also summarized in Alg.~\ref{alg:eb_train}; the full pseudocode is detailed in Appendix~\ref{app:lbs_train_proc}.

\begin{figure}[t]
\begin{minipage}[t]{0.48\textwidth}
\begin{algorithm}[H]
\small
\caption{EB-Decode Training}
\label{alg:eb_train}
\begin{algorithmic}[1]
\Require Frozen dLLM $\mathcal{F}_\theta$, prompt-response pair $\mathcal{D}$
\Statex \smash{\rlap{\hspace*{-1.5em}\textcolor{lightred}{\rule[-5.3\baselineskip]{\dimexpr\linewidth+2em\relax}{6\baselineskip}}}}\textcolor{lbsfg}{\textit{// LBS router $\mathcal{R}_\phi$}}
\For{$x$ $\in \mathcal{D}$}
  \State $\tilde{H}^i, \hat{x}^i \gets \mathcal{F}_\theta(\text{Mask}(x))$ \Comment{LBS Inputs}
  \State $O^i_\phi \gets \mathcal{R}_\phi(\tilde{H}^i, \hat{x}^i)$
  \State Update $\phi$ via $\mathcal{L}_{\mathrm{LBS}}$ \Comment{Eq.~\ed{\eqref{eq:lbs_loss}}}
\EndFor
\Statex \smash{\rlap{\hspace*{-1.5em}\textcolor{lightblue}{\rule[-7.3\baselineskip]{\dimexpr\linewidth+2em\relax}{7.95\baselineskip}}}}\textcolor{lpsfg}{\textit{// LPS router $\mathcal{R}_\psi$}}
\For{$x \in \mathcal{D}$}
  \State Decode with LBS; record trace $\mathcal{T}$ %
\EndFor
\For{$(f^i, \mathrm{pos}^i) \subset \mathcal{T}$} \Comment{LPS Inputs}
  \State $O^i_\psi \gets \mathcal{R}_\psi(f^i, \mathrm{pos}^i)$  
  \State Update $\psi$ via $\mathcal{L}_{\mathrm{LPS}} $\Comment{Eq.~\ed{\eqref{eq:lps_loss}}}
  \EndFor
\State \Return $\mathcal{R}_\phi, \mathcal{R}_\psi$ %
\end{algorithmic}
\end{algorithm}

\end{minipage}\hfill
\begin{minipage}[t]{0.48\textwidth}
\begin{algorithm}[H]
\small
\caption{EB-Decode Inference}
\label{alg:eb_infer}
\begin{algorithmic}[1]
\Require $\mathcal{F}_\theta$, $\mathcal{R}_\phi$, $\mathcal{R}_\psi$, and thresholds $\tau$, $\tau_\psi$
\State $x \gets \text{Prompt } \& \texttt{[MASK]}$
\While{any $x^i = \texttt{[MASK]}$}
  \State $\tilde{H}^i, \hat{x}^i, f^i \gets \mathcal{F}_\theta(x)$ %
  \State $f^i = (\kappa^i, \tilde{H}^i, \Delta^i, \rho)$ \Comment{Feature Details}
  \Statex \smash{\rlap{\hspace*{-1.5em}\textcolor{lightred}{\rule[-3.1\baselineskip]{\dimexpr\linewidth+2em\relax}{3.75\baselineskip}}}}\textcolor{lbsfg}{\textit{\quad\,\,\,\,// LBS}}
  \If{no active block}
    \State $\mathcal{B} \gets \mathrm{Filters}\bigl(\mathcal{R}_\phi(\tilde{H}^i,\hat{x}^i)\bigr)$
  \EndIf
  \Statex \smash{\rlap{\hspace*{-1.7em}\textcolor{lightblue}{\rule[-3.6\baselineskip]{\dimexpr\linewidth+2em\relax}{4.3\baselineskip}}}}\textcolor{lpsfg}{\textit{\quad\,\,\,\,// LPS}}
  \State $O^i_\psi \gets \mathcal{R}_\psi(f^i, \mathrm{pos}^i)$
  \State $\mathcal{A} \gets \{i \in \mathcal{B}: \kappa^i \ge \tau \,\vee\, \pp{p^i_\psi} > \tau_\psi\}$ %
  \State $x^i \gets \hat{x}^i$ for $i \in \mathcal{A}$ \Comment{Accept \& Commit}
  \State If $\mathcal{B}$ resolved, move to Line 5
\EndWhile
\State \Return $x$
\end{algorithmic}
\end{algorithm}
\end{minipage}
\end{figure}

\textbf{LBS Router Inference.}
As shown in Alg.~\ref{alg:eb_infer},
at each step, entropy $\tilde{H}^i$ and token ID $\hat{x}^i$ are computed from the dLLM output logits over masked positions and passed to $\mathcal{R}_\phi$, producing a candidate block $\mathcal{C} = \{i : p^i_\phi > 0.5\}$.
This candidate block is then refined by three filters, each addressing a specific failure mode.
(1) \emph{EOS removal}.
End-of-sequence tokens have near-zero entropy, so the router nearly always selects them. However, unmasking EOS tokens too early in the inference process would truncate generation prematurely and severely damages accuracy on tasks that require step-by-step reasoning like math and coding \citep{huang2025pcsampler}. Thus, we remove all EOS tokens from $\mathcal{C}$ entirely. We quantify the dominance of EOS predictions at the sequence tail in Appendix~\ref{app:eoszone}.
(2) \emph{Max-gap constraint} ($g_{\max}$).
Ideally, we want to identify a homogeneous difficulty plateau as the block, instead of cherry-picking isolated easy tokens scattered across a hard region.
If consecutive selected positions are separated by more than $g_{\max}$ tokens, the candidate block straddles an unresolved hard region, violating the spatial coherence assumption and forcing the model to denoise positions that have not yet been committed.
Thus, starting from the leftmost selected position, any candidate whose gap to its predecessor exceeds $g_{\max}$ is discarded along with all subsequent selected positions.
(3) \emph{Min-tokens constraint} ($L_{\min}$).
After applying the above two filters, if the resulting candidate set contains fewer than $L_{\min}$ tokens, LBS falls back to selecting the first 32 masked positions in left-to-right order. 
This constraint is critical for accuracy. If the number of tokens is too small after filtering, the router can degenerate to selecting scattered easy tokens anywhere in the sequence. 
In summary, the router makes per-token decisions, resulting in an inherently non-contiguous candidate block that adapts to the entropy pattern; contiguity is instead imposed at inference time via filtering.

\subsection{Enabler 2: Learnable Parallel Sampling (LPS)}
\label{sec:LPS}

LBS adapts block boundaries to local difficulty but keeps the commit rule fixed, leaving intra-block parallelism underutilized.
The early convergence observation in Sec.~\ref{sec:observation of EB Phenomenon in dLLM Decoding} shows that a masked position\ed{'}s top-1 prediction often becomes correct several steps before its confidence reaches the threshold, making additional refinement steps unnecessary.
To bridge this gap, we introduce LPS, a lightweight router $\mathcal{R}_\psi$ that learns when a masked position within a variable-length block is ready to be finalized, enabling early commitment before reaching the confidence threshold, as shown in Fig.~\ref{fig:overview} (c).

\textbf{Position-aware LPS Router.}  
Unlike many existing decoding methods~\citep{nie2025large,ye2025dream,wu2025fast,lu2025adablock} that rely on top-1 confidence $\kappa^i$ as a static commit criterion, LPS learns this decision, using confidence as the primary input and augmenting it with two signals.
First, to make the LPS router position-aware, we incorporate the token's normalized intra-block position embedding,
$\mathrm{pos}^i = \sfrac{(i-b)}{|\mathcal{B}|} \in [0,1]$, where $b$ is the left boundary index and $|\mathcal{B}|$ is the block size. This encoding captures the token's relative location within the block and enables the router to correlate decisions across positions rather than scoring them independently.
To support variable-length block inputs, our LPS router adopts a two-layer Transformer architecture to leverage positional information, rather than an MLP-based design~\citep{bao2025learning} that cannot handle variable-length inputs. Model details are provided in Appendix~\ref{app:lps_arch}.
Second, to enable the LPS router to distinguish correct-but-underconfident tokens from genuinely uncertain ones, we incorporate three auxiliary uncertainty features:
(1) the normalized entropy \ed{$\tilde{H}^i_t$} for absolute prediction uncertainty\new{, which alone cannot tell whether the top-1 candidate dominates};
(2) the confidence gap $\Delta^i_t$ between the top-1 and top-2 predictions for relative label ambiguity\new{, i.e., the margin over the strongest competitor}; and
(3) the block-wise mask ratio $\rho_t$ to capture global decoding progress\new{, as the commit policy should differ across denoising stages}.
Overall, for each masked position $i \in \mathcal{M}$ in the active block $\mathcal{B}$, the input to the LPS router is defined as the concatenation of all features \ed{$f^i_t = (\kappa^i_t, \tilde{H}^i_t, \Delta^i_t, \rho_t)$}, and the output is \ed{$O^i_\psi = \mathcal{R}_\psi(f^i_t, \mathrm{pos}^i)$}\pp{, from which the commit probability is obtained as $p^i_\psi = \sigma(O^i_\psi)$}.

\textbf{Self-supervised LPS Router Training.} We train LPS using a \textit{generate-then-replay} pipeline that requires no human label. We first run the base dLLM under its default decoding schedule and record the resulting completed sequence $x_0$.
We treat $x_0$ as a convergence target rather than absolute ground truth, so \ed{the} LPS router learns to predict when a token has stabilized to the model\ed{'}s own final prediction.
During replay, at each intermediate step $t$, we recompute the features $f^i_t$ and assign $y^i_t = \mathbf{1}[\hat{x}^i_t = x^i_0]$, indicating whether the current top-1 prediction already matches $x_0$.
To keep the replay aligned with the original generation trajectory, we apply an oracle rule that unmasks positions where the current prediction matches $x_0$, and replaces mismatched positions with their corresponding tokens from $x_0$.
Since prematurely committing an incorrect token is irreversible, we use a weighted binary cross-entropy loss that penalizes false positives more heavily than false negatives during training:
\begin{equation}
    \mathcal{L}_{\mathrm{LPS}} = \frac{1}{|\mathcal{M}_t|} \sum_{i \in \mathcal{M}_t} \pp{w^i_t} \cdot \mathrm{BCE}\Bigl(\sigma\bigl( \mathcal{R}_\psi(f^i_t, \mathrm{pos}^i_t) \bigr), y^i_t\Bigr),
\label{eq:lps_loss}
\end{equation}
where $\sigma$ indicates a sigmoid function, negative samples (incorrect early commitments) receive weights $w_t^i > 1$, causing false positives to dominate the gradient.
The training procedure is briefly summarized in Alg.~\ref{alg:eb_train}, with full trace-generation and training details provided in Appendix~\ref{app:lps_trace} and \ref{app:lps_training}.

\textbf{LPS Router Inference.} 
As shown in Alg.~\ref{alg:eb_infer},
at each decoding step, the base dLLM produces top-1 predictions and confidences for all masked positions. If no active block exists, LBS selects a new $\mathcal{B}$; otherwise, the previous $\mathcal{B}$ is reused.
Within $\mathcal{B}$, a position $i$ is committed if its base confidence exceeds the threshold $\tau$ or its LPS score exceeds a threshold $\tau_\psi$ tuned on a held-out validation split.
In this way, LPS preserves the standard threshold rule for tokens the model is already confident about, while committing correct-but-underconfident tokens that would otherwise be deferred.
If no position in $\mathcal{B}$ satisfies either condition, we fall back to committing the highest-confidence masked position to prevent decoding from stalling, following standard practice in confidence-based decoding.
Once $\mathcal{B}$ is fully resolved, decoding proceeds to the next block. 
Full pseudocode is provided in Appendix~\ref{app:eb_infer_proc}.

\section{Experiments}
\label{sec:experiments}

\subsection{Experimental Setup}
\label{sec:exp_setup}

\textbf{Models, Benchmarks, and Metrics.}
\textit{Models.}
We evaluate EB-Decode on three open-source dLLMs: LLaDA-8B-Instruct~\citep{nie2025large}, Dream-v0-Instruct-7B~\citep{ye2025dream}, and LLaDA-1.5~\citep{zhu2025llada}.
\textit{Benchmarks.}
Our evaluation covers mathematical reasoning on GSM8K~\citep{cobbe2021training} (0-shot) and MATH500~\citep{lightman2023lets} (0-shot), as well as code generation on HumanEval~\citep{chen2021evaluating} (0-shot) and MBPP~\citep{austin2021program} (3-shot). For GSM8K and MATH500, we append a chain-of-thought prompt suffix~\citep{wei2022chain} to each question following~\citep{chen2025dparallel}.
\textit{Metrics.}
We report task accuracy, throughput, and the associated speedups compared to the base models. Detailed experimental settings are provided in Appendix~\ref{app:exp_setting}.

\begin{table}[!t]
\centering
\caption{Comparison of EB-Decode and baseline methods across four benchmarks and three base models. We report tokens per second per GPU (TPS), speedup (Sp.up), and accuracy (Acc.\%). Speedup is measured relative to the vanilla decoding method. The best throughput and speedup are highlighted in \textbf{bold}. ``---'' indicates not applicable, as Learn2PD \ed{provides no open-source implementation for} LLaDA-1.5.}
\label{tab:main_results}
\setlength{\tabcolsep}{3pt}
\small
\begin{tabular}{l ccc | ccc | ccc}
\toprule
& \multicolumn{3}{c}{\textbf{LLaDA-8B-Instruct}} & \multicolumn{3}{c}{\textbf{Dream-v0-Instruct-7B}} & \multicolumn{3}{c}{\textbf{LLaDA-1.5}} \\
\cmidrule(lr){2-4} \cmidrule(lr){5-7} \cmidrule(lr){8-10}
\textbf{Method} & \textbf{Sp.up} & \textbf{TPS} & \textbf{Acc.\,(\%)} & \textbf{Sp.up} & \textbf{TPS} & \textbf{Acc.\,(\%)} & \textbf{Sp.up} & \textbf{TPS} & \textbf{Acc.\,(\%)} \\
\midrule
\multicolumn{10}{c}{\textit{HumanEval}} \\
\cmidrule(lr){1-10}
Vanilla              & 1.00$\times$         & 23.70           & 43.90 & 1.00$\times$          & 11.33           & 54.88 & 1.00$\times$          & 23.79           & 43.90 \\
Fast-dLLM~\citep{wu2025fast}            & 2.66$\times$         & 63.08           & 43.90 & 4.29$\times$          & 48.57           & 53.66 & 2.72$\times$          & 64.60           & 42.68 \\
AdaBlock-dLLM~\citep{lu2025adablock}        & 2.38$\times$         & 56.33           & 42.68 & 4.05$\times$          & 45.88           & 52.44 & 2.52$\times$          & 59.95           & 41.46 \\
Learn2PD~\citep{bao2025learning}                  & 2.78$\times$         & 65.92           & 42.68 & 2.15$\times$          & 24.31           & 53.66 & ---                   & ---             & ---   \\
\rowcolor[RGB]{235,247,235} EB-Decode (LBS) & 2.90$\times$         & 68.83           & 45.73 & 4.46$\times$          & 50.55           & 55.49 & 2.68$\times$          & 63.73           & 41.46 \\
\rowcolor[RGB]{235,243,252} EB-Decode (LBS+LPS)     & \textbf{3.53$\times$} & \textbf{83.69}  & 43.29 & \textbf{5.61$\times$} & \textbf{63.53}  & 54.88 & \textbf{3.58$\times$} & \textbf{85.19}  & 42.68 \\
\midrule
\multicolumn{10}{c}{\textit{MBPP}} \\
\cmidrule(lr){1-10}
Vanilla              & 1.00$\times$         & 11.95           & 40.00 & 1.00$\times$          & 2.23            & 53.40 & 1.00$\times$          & 8.28            & 40.80 \\
Fast-dLLM~\citep{wu2025fast}            & 4.13$\times$         & 49.31           & 40.20 & 11.88$\times$         & 26.49           & 53.00 & 6.41$\times$          & 53.13           & 40.20 \\
AdaBlock-dLLM~\citep{lu2025adablock}        & 4.02$\times$         & 48.02           & 41.20 & 11.54$\times$         & 25.74           & 52.40 & 6.01$\times$          & 49.82           & 40.60 \\
Learn2PD~\citep{bao2025learning}                  & 4.72$\times$         & 56.37           & 40.00 & 4.04$\times$          & 9.01            & 52.00 & ---                   & ---             & ---   \\
\rowcolor[RGB]{235,247,235} EB-Decode (LBS) & 4.47$\times$         & 53.37           & 40.40 & 14.70$\times$         & 32.77           & 54.80 & 6.39$\times$          & 52.94           & 40.60 \\
\rowcolor[RGB]{235,243,252} EB-Decode (LBS+LPS)     & \textbf{5.37$\times$} & \textbf{64.18}  & 40.20 & \textbf{18.76$\times$} & \textbf{41.83}  & 53.60 & \textbf{7.76$\times$} & \textbf{64.26}  & 39.80 \\
\midrule
\multicolumn{10}{c}{\textit{GSM8K-CoT}} \\
\cmidrule(lr){1-10}
Vanilla              & 1.00$\times$         & 17.04           & 79.07 & 1.00$\times$          & 12.03           & 77.78 & 1.00$\times$          & 16.94           & 79.76 \\
Fast-dLLM~\citep{wu2025fast}            & 4.65$\times$         & 79.20           & 78.70 & 4.51$\times$          & 54.20           & 77.86 & 5.15$\times$          & 87.27           & 79.45 \\
AdaBlock-dLLM~\citep{lu2025adablock}        & 4.18$\times$         & 71.22           & 79.00 & 4.57$\times$          & 54.97           & 78.70 & 4.89$\times$          & 82.87           & 79.00 \\
Learn2PD~\citep{bao2025learning}                  & 4.91$\times$         & 83.64           & 78.24 & 2.07$\times$          & 24.96           & 78.09 & ---                   & ---             & ---   \\
\rowcolor[RGB]{235,247,235} EB-Decode (LBS) & 5.09$\times$         & 86.80           & 79.68 & 4.92$\times$          & 59.22           & 78.10 & 4.79$\times$          & 81.20           & 80.36 \\
\rowcolor[RGB]{235,243,252} EB-Decode (LBS+LPS)     & \textbf{6.14$\times$} & \textbf{104.55} & 78.24 & \textbf{6.36$\times$} & \textbf{76.47}  & 77.33 & \textbf{6.25$\times$} & \textbf{105.91} & 79.53 \\
\midrule
\multicolumn{10}{c}{\textit{MATH500-CoT}} \\
\cmidrule(lr){1-10}
Vanilla              & 1.00$\times$         & 21.72           & 42.40 & 1.00$\times$          & 24.97           & 51.00 & 1.00$\times$          & 15.10           & 43.20 \\
Fast-dLLM~\citep{wu2025fast}            & 3.26$\times$         & 70.77           & 42.00 & 3.10$\times$          & 77.39           & 51.40 & 5.00$\times$          & 75.59           & 42.80 \\
AdaBlock-dLLM~\citep{lu2025adablock}        & 3.07$\times$         & 66.63           & 41.60 & 3.08$\times$          & 77.00           & 48.80 & 4.69$\times$          & 70.88           & 42.20 \\
Learn2PD~\citep{bao2025learning}                  & 3.78$\times$         & 82.11           & 41.80 & 1.12$\times$          & 28.05           & 49.80 & ---                   & ---             & ---   \\
\rowcolor[RGB]{235,247,235} EB-Decode (LBS) & 3.32$\times$         & 72.03           & 41.20 & 2.96$\times$          & 73.91           & 49.80 & 4.92$\times$          & 74.24           & 42.60 \\
\rowcolor[RGB]{235,243,252} EB-Decode (LBS+LPS)     & \textbf{3.86$\times$} & \textbf{83.79}  & 42.40 & \textbf{3.57$\times$} & \textbf{89.06}  & 49.40 & \textbf{5.55$\times$} & \textbf{83.79}  & 42.40 \\
\bottomrule
\end{tabular}
\end{table}

\textbf{Baselines.}
We compare EB-Decode against four inference-stage baselines: (1) the vanilla block-wise decoding method used by each base dLLM~\citep{nie2025large,ye2025dream,zhu2025llada}, which serves as the no-acceleration reference; (2) Fast-dLLM~\citep{wu2025fast}, which commits multiple tokens per step based on a confidence threshold \new{(for a fair comparison, we use its parallel decoding without KV cache, and report its results with KV cache in Tab.~\ref{tab:abl_cache})}; (3) AdaBlock-dLLM~\citep{lu2025adablock}, which adaptively places block boundaries using delimiter-token confidence; and (4) Learn2PD~\citep{bao2025learning}, which employs a lightweight filter for parallel decoding.
\new{A detailed comparison between EB-Decode and these and related methods is given in Appendix~\ref{app:positioning}.}

\textbf{Router Training Details.}
\new{Both routers are lightweight, with about $0.6$M trainable parameters for LBS and $0.03$M for LPS.}
The LBS router is trained on dLLM-generated responses from $92{,}000$ prompts randomly sampled from GSM8K~\citep{cobbe2021training}, the PRM12K training set~\citep{lightman2023lets}, and a subset of the Numina-Math dataset~\citep{numina_math_datasets}, following the setup of~\citep{chen2025dparallel}. The router is trained for 6 epochs using AdamW with a learning rate of $5\mathrm{e}{-5}$. The actual training of the LBS router takes only about $2$ hours.
The LPS router is trained on $10{,}000$ prompts randomly sampled from AQUA-RAT~\citep{ling2017program}. Training uses AdamW with a learning rate of $1\mathrm{e}{-3}$ and is early-stopped when the validation recall does not improve for $100$ \ed{epochs}. Trace generation takes about $2$ hours, while router training itself requires only around $10$ minutes. Further details are provided in Appendix~\ref{app:lbs_setup} and~\ref{app:lps_setup}.

\subsection{Comparison of EB-Decode with SOTA Baselines}
\label{sec:main_results}

Tab.~\ref{tab:main_results} summarizes the comparison between our proposed EB-Decode and four baselines in terms of accuracy and throughput across three dLLMs and four benchmarks.
We observe that both EB-Decode with LBS only and the full EB-Decode with LBS+LPS consistently achieve better accuracy-efficiency trade-offs.
EB-Decode with LBS alone already improves throughput across all settings, while even improving accuracy in several cases. 
For example, \ed{on LLaDA-8B-Instruct with HumanEval}, LBS increases throughput from 23.70 to 68.83 TPS while simultaneously improving accuracy from 43.90\% to 45.73\%. 
This result suggests that adaptive block boundaries not only unlock greater parallelism, but also provide difficult tokens with richer right-context information during decoding.
The full EB-Decode improves throughput by \textbf{\ed{3.53--18.76$\times$}} over the vanilla decoder, while maintaining comparable accuracy (around $\pm1\%$). \ed{EB-Decode also outperforms the strongest baseline in every setting, delivering up to 1.58$\times$ higher throughput than Fast-dLLM while maintaining comparable or even higher accuracy.}
\ed{These results demonstrate} the effectiveness of our EB-Decode.

\subsection{Ablation Studies of EB-Decode}
\label{sec:ablation}

\begin{table}[t]
\begin{minipage}[t]{0.45\textwidth}
\centering
\caption{Ablation of LBS components on LLaDA-8B-Instruct and GSM8K-CoT. Each row progressively adds one component.}
\label{tab:abl_lbs}
\setlength{\tabcolsep}{3pt}
\small
\begin{tabular}{l|cc}
\toprule
\textbf{Variant} & \textbf{Acc.\,(\%)} & \textbf{TPS} \\
\midrule
LBS Router                                  & 74.53          & 42.25          \\
\midrule
+ EOS removal                               & 77.33          & 90.31          \\
+ Max-gap constraint                        & 78.70          & 86.92          \\
+ Min-tokens constraint & 79.68 & 86.80 \\
\bottomrule
\end{tabular}
\end{minipage}\hfill
\begin{minipage}[t]{0.52\textwidth}
\centering
\caption{Ablation of LPS components on LLaDA-8B-Instruct and GSM8K-CoT. Each row progressively adds one component.}
\label{tab:abl_lps}
\setlength{\tabcolsep}{3pt}
\small
\begin{tabular}{l|cc}
\toprule
\textbf{Variant} & \textbf{Acc.\,(\%)} & \textbf{TPS} \\
\midrule
LPS Router in MLP            & 78.39 & 87.40           \\
\midrule
+ MLP $\rightarrow$ Transformer    & 78.25          & 97.31           \\
+ Intra-block position          & 77.94          & 104.18          \\
+ Entropy + Gap + Mask Ratio & 78.24 & 104.55 \\
\bottomrule
\end{tabular}
\end{minipage}
\end{table}

\begin{wraptable}[13]{r}{0.5\textwidth}
\centering
\begingroup
\vspace{-2em}
\caption{Accuracy and TPS on LLaDA-8B-Instruct and GSM8K-CoT with the three filters applied progressively without the LBS router, where blocks are selected randomly. Even with all three filters, both accuracy and TPS remain below those of LBS.}
\label{tab:selector_control}
\setlength{\tabcolsep}{3pt}
\small
\begin{tabular}{l|cc}
\toprule
\textbf{Variant} & \textbf{Acc.\,(\%)} & \textbf{TPS} \\
\midrule
w/o LBS router            & 49.13 & 11.56 \\
\midrule
+ EOS removal             & 68.76 & 24.24 \\
+ Max-gap constraint      & 76.27 & 23.16 \\
+ Min-tokens constraint   & 76.80 & 23.96 \\
\midrule
\rowcolor[RGB]{235,247,235} LBS router + all filters (ours) & \textbf{79.68} & \textbf{86.80} \\
\bottomrule
\end{tabular}
\vspace{-3cm}
\endgroup
\end{wraptable}

\textbf{LBS Design Choices.}
Tab.~\ref{tab:abl_lbs} additively builds up the LBS filter described in Sec.~\ref{sec:adablock}. Starting from the LBS router's selected block, we incrementally add EOS removal, the max-gap constraint, and the min-tokens constraint to further refine the selected block. The results in the table empirically demonstrate that all filters are crucial in maximizing the task accuracy while maintaining high TPS. \new{To check whether these gains stem from the filters alone, Tab.~\ref{tab:selector_control} removes the LBS router while keeping the same filters. Even with all three filters, this variant reaches only $76.80\%$ accuracy at $23.96$ TPS, $2.88$ points lower and $3.62\times$ slower than LBS with the same filters, showing that the learned block selection itself brings large gains.} \ed{We also train an LBS router of comparable size on LLaDA-8B-Instruct that takes only normalized entropy as input. Its converged training loss is higher, and accuracy across the four benchmarks drops by 1.2 points on average at similar TPS.} This suggests that semantic information can indeed help the LBS router pick a more optimal block.

\textbf{LPS Design Choices.}
Tab.~\ref{tab:abl_lps} additively builds up the LPS router described in Sec.~\ref{sec:LPS}, on top of LBS selected blocks. Starting from \ed{an} MLP taking only confidence as input, we replace the encoder with a small Transformer, add the intra-block position embedding, and add the three auxiliary uncertainty features (entropy, top-1/top-2 gap, mask ratio). The results in the table empirically demonstrate the benefit of using a Transformer architecture over an MLP, and show that all components contribute to maximizing the TPS while preserving the task accuracy.

\begin{wraptable}{r}{0.5\textwidth}
\centering
\vspace{-0.5em}
\caption{Accuracy and speedup on LLaDA-8B-Instruct and GSM8K-CoT at different generation lengths. EB-Decode delivers higher \ed{speedup} at both lengths with accuracy comparable to \ed{the vanilla decoder}.}
\vspace{0.1cm}
\label{tab:abl_genlen}
\setlength{\tabcolsep}{3pt}
\small
\begin{tabular}{l | cc | cc}
\toprule
& \multicolumn{2}{c|}{$L{=}256$} & \multicolumn{2}{c}{$L{=}512$} \\
\cmidrule(lr){2-3} \cmidrule(lr){4-5}
\textbf{Method} & \textbf{Acc.\,(\%)} & \textbf{Sp.up} & \textbf{Acc.\,(\%)} & \textbf{Sp.up} \\
\midrule
Vanilla            & 75.13 & 1.00$\times$           & 79.07 & 1.00$\times$               \\
\rowcolor[RGB]{235,247,235} LBS     & 74.98 & 3.27$\times$       & 79.68 & 5.09$\times$               \\
\rowcolor[RGB]{235,243,252} LBS+LPS & 74.60 & \textbf{3.86$\times$}   & 78.24 & \textbf{6.14$\times$}     \\
\bottomrule
\end{tabular}
\end{wraptable}
\textbf{Performance on Different Generation Lengths.}
Tab.~\ref{tab:abl_genlen} compares EB-Decode against the baseline at generation lengths $L\in\{256, 512\}$. EB-Decode achieves the highest throughput and comparable accuracy at both lengths. Also, the speedup grows as the generation length increases, indicating that EB-Decode may be especially effective for long-sequence generation.

\begin{wraptable}{r}{0.5\textwidth}
\centering
\vspace{-1em}
\caption{Accuracy and TPS on LLaDA-8B-Instruct and GSM8K-CoT across different cache strategies\new{, for both EB-Decode and Fast-dLLM}.}
\label{tab:abl_cache}
\setlength{\tabcolsep}{3pt}
\vspace{0.1cm}
\small
\begin{tabular}{lcc}
\toprule
\textbf{Method ($L = 512$)} & \textbf{Acc.\,(\%)} & \textbf{TPS} \\
\midrule
\new{Fast-dLLM~\citep{wu2025fast}}          & \new{78.70}    & \new{50.90}     \\
\new{Fast-dLLM (Prefix Cache)}              & \new{78.92}    & \new{62.03}     \\
\new{Fast-dLLM (Dual Cache)}                & \new{78.70}    & \new{56.92}     \\
\midrule
EB-Decode                                  & \textbf{78.24} & 65.27           \\
EB-Decode (Prefix Cache)                   & 77.93          & 75.95           \\
EB-Decode (Dual Cache) & 77.93 & \textbf{78.43}  \\
\bottomrule
\end{tabular}
\vspace{-2em}
\end{wraptable}

\textbf{Effect of KV-Cache Configuration.}
Tab.~\ref{tab:abl_cache} shows the result of using different KV-cache strategies on top of EB-Decode on 4$\times$A100. Since LBS uses non-contiguous active blocks, we apply cache at the block boundaries: prefix cache~\citep{wu2025fast} reuses keys/values for tokens to the left of the leftmost selected position, and dual cache~\citep{wu2025fast} additionally reuses keys/values for tokens to the right of the rightmost selected position. Both variants further improve the throughput of EB-Decode while preserving accuracy, demonstrating that EB-Decode can integrate with standard KV-cache pipelines for additional speedup.
\new{With KV caching being considered on top of both methods, EB-Decode still outperforms Fast-dLLM by $1.28\times$, $1.22\times$, and $1.38\times$ without cache, with the prefix cache, and with the dual cache, respectively, showing that its advantage holds even when the baseline also uses KV caching.}

\textbf{Overhead of LBS and LPS Routers.}
We measure the wall-clock overhead of the two routers as a fraction of the per-step backbone forward time on A100. LBS accounts for roughly $3.80\%$ and LPS for $2.41\%$ of the per-step latency (6.21\% in total), confirming that the speedups come from jointly decoding low-entropy tokens earlier before they reach the confidence threshold and thereby using fewer decoding steps, while the routers themselves introduce only negligible inference overhead.

\subsection{\texorpdfstring{\new{Generality and Broader Comparisons}}{Generality and Broader Comparisons}}
\label{sec:generality}

\begin{table}[t]
\centering
\caption{Router transfer without retraining on GSM8K-CoT, (a) across models and (b) across generation lengths. ``Native'' uses routers trained in the target setting, and ``Transferred'' uses the router(s) listed in the second column.}
\label{tab:transfer}
\setlength{\tabcolsep}{4pt}
\small
\begin{tabular}{ll|cc|cc}
\toprule
& & \multicolumn{2}{c|}{\textbf{Native}} & \multicolumn{2}{c}{\textbf{Transferred}} \\
\cmidrule(lr){3-4} \cmidrule(lr){5-6}
\textbf{Target} & \textbf{Transferred router} & \textbf{Acc.\,(\%)} & \textbf{TPS} & \textbf{Acc.\,(\%)} & \textbf{TPS} \\
\midrule
\multicolumn{6}{c}{\textit{(a) Across models}} \\
\midrule
\mg{LLaDA-1.5}            & \mg{LBS from LLaDA-8B}        & \mg{79.23} & \mg{64.95} & \mg{79.15} & \mg{64.68} \\
LLaDA-1.5             & LBS + LPS from LLaDA-8B   & 79.23 & 64.95 & 79.23 & 66.37 \\
Dream-v0-Instruct-7B  & LPS from LLaDA-8B         & 76.19 & 43.55 & 76.49 & 43.07 \\
\midrule
\multicolumn{6}{c}{\textit{(b) Across generation lengths (LLaDA-8B-Instruct)}} \\
\midrule
$L{=}512$             & LPS trained at $L{=}256$  & 78.17 & 69.54 & 78.24 & 66.95 \\
$L{=}256$             & LPS trained at $L{=}512$  & 74.68 & 95.37 & 74.15 & 94.08 \\
\bottomrule
\end{tabular}
\end{table}

\new{\textbf{Router Transferability.} The routers of EB-Decode are largely transferable, so they can be reused without retraining when the deployment setting changes. We examine this transferability from three aspects: across models, across generation lengths, and across tasks. (1) Across models as shown in Tab.~\ref{tab:transfer} (a), \ed{we consider two transfer targets. For LLaDA-1.5, which belongs to the same model series as LLaDA-8B-Instruct, transferring LBS alone or together with LPS matches the natively trained routers in accuracy and throughput.} \ed{For Dream-v0-Instruct-7B, which comes from a different model series, LBS does not transfer, because it embeds predicted token IDs with the base model's embedding table and vocabularies differ across series. LPS, whose features are vocabulary-independent, still transfers.} (2) Across generation lengths as shown in Tab.~\ref{tab:transfer} (b), an LPS router trained at one length remains on par at the other. \mg{\ed{Only LPS is evaluated here, because} LBS takes the entire masked sequence as input and uses positional embeddings tied to the generation length (Appendix~\ref{app:router_arch})\ed{. We therefore recommend using LBS} at its training length.} (3) Across tasks, both routers are trained only on mathematical data, so the HumanEval and MBPP results in Tab.~\ref{tab:main_results} already reflect transfer to unseen code tasks.}

\begin{table}[t]
\centering
\caption{Accuracy and throughput of LBS on block-forward dLLMs, which compute logits only for the current active window. ``Thr.'' denotes the commit threshold; accuracy gains over the baseline are shown in parentheses.}
\label{tab:blockforward}
\setlength{\tabcolsep}{4pt}
\small
\begin{tabular}{l c | cc | cc}
\toprule
& & \multicolumn{2}{c|}{\textbf{HumanEval (0-shot)}} & \multicolumn{2}{c}{\textbf{GSM8K (5-shot)}} \\
\cmidrule(lr){3-4} \cmidrule(lr){5-6}
\textbf{Method} & \textbf{Thr.} & \textbf{Acc.\,(\%)} & \textbf{Token/s} & \textbf{Acc.\,(\%)} & \textbf{Token/s} \\
\midrule
\multicolumn{6}{c}{\textit{LLaDA 2.0~\citep{bie2025llada2}, active window $64$}} \\
\cmidrule(lr){1-6}
LLaDA 2.0                                        & 0.8  & 75.6          & 12.9 & 91.1          & 11.7 \\
\rowcolor[RGB]{235,247,235} EB-Decode (LBS)      & 0.8  & \textbf{77.4} \textit{(+1.8)} & 12.0 & \textbf{91.6} \textit{(+0.5)} & 11.1 \\
LLaDA 2.0                                        & 0.95 & 79.9          & 9.4  & 91.8          & 8.6  \\
\rowcolor[RGB]{235,247,235} EB-Decode (LBS)      & 0.95 & \textbf{82.3} \textit{(+2.4)} & 8.8  & \textbf{92.3} \textit{(+0.5)} & 8.2  \\
\midrule
\multicolumn{6}{c}{\textit{SDAR-1.7B-Chat-b32~\citep{cheng2025sdar}, active window $32$}} \\
\cmidrule(lr){1-6}
SDAR                                             & 0.8  & 42.1          & 38.4 & 68.5          & 39.3 \\
\rowcolor[RGB]{235,247,235} EB-Decode (LBS)      & 0.8  & \textbf{43.9} \textit{(+1.8)} & 33.8 & \textbf{69.5} \textit{(+1.0)} & 35.8 \\
SDAR                                             & 0.9  & 46.3          & 34.2 & 73.1          & 32.8 \\
\rowcolor[RGB]{235,247,235} EB-Decode (LBS)      & 0.9  & \textbf{47.6} \textit{(+1.3)} & 30.2 & \textbf{73.6} \textit{(+0.5)} & 29.9 \\
\bottomrule
\end{tabular}
\end{table}

\begin{table}[t]
\centering
\caption{Comparison with AR models on GSM8K-CoT ($L{=}512$). Acc.\ follows the standard answer extraction of the GSM8K harness, while Flex.\ Acc.\ takes the content of \texttt{\textbackslash boxed\{\}} or the last number in the response and is thus robust to the output format.}
\label{tab:ar_baseline}
\setlength{\tabcolsep}{4pt}
\small
\begin{tabular}{ll ccc}
\toprule
\textbf{Model} & \textbf{Type} & \textbf{TPS} & \textbf{Acc.\,(\%)} & \textbf{Flex.\ Acc.\,(\%)} \\
\midrule
Qwen3-8B~\citep{qwen3}                            & AR   & 21.22          & 32.22 & \textbf{91.89} \\
LLaMA3-8B-Instruct~\citep{grattafiori2024llama}                 & AR   & 28.94          & 68.54 & 80.36 \\
\midrule
LLaDA-8B-Instruct, Vanilla~\citep{nie2025large}   & dLLM & 10.24          & \textbf{79.07} & 84.08 \\
\rowcolor[RGB]{235,243,252} LLaDA-8B-Instruct, EB-Decode & dLLM & \textbf{69.54} & 78.17 & 82.49 \\
\bottomrule
\end{tabular}
\end{table}

\new{\textbf{Compatibility with Block-Forward dLLMs.} Unlike the bidirectional dLLMs in Tab.~\ref{tab:main_results}, which forward the full sequence at every step, block-forward dLLMs such as LLaDA 2.0~\citep{bie2025llada2} and SDAR~\citep{cheng2025sdar} adopt block-causal attention and compute logits only for the current active window. \ed{This does not prevent EB-Decode from applying. LBS selects the next easy-to-decode block, so its selections concentrate near the decoding frontier, which the active window already covers. Adapting to block-causal attention then requires a single parameter change, tightening $g_{\max}$ from $5$ to $1$ so that each block is filled contiguously. Restricted to the window logits, LBS improves accuracy in all eight configurations of Tab.~\ref{tab:blockforward}, at both commit thresholds tested. Throughput stays close to the baseline, since LBS reuses logits the engine already computes and adds no extra forward. LPS needs no adaptation, since its input covers only the tokens inside the selected block.}}

\new{\textbf{Comparison with Autoregressive Models.} To examine whether EB-Decode narrows the efficiency gap between dLLMs and autoregressive (AR) models, Tab.~\ref{tab:ar_baseline} compares it with Qwen3-8B~\citep{qwen3} and LLaMA3-8B-Instruct~\citep{grattafiori2024llama}, with all models run on the same \texttt{transformers} backend with batch size $1$ on $4\times$A100 GPUs. \mg{The low Acc.\ of the AR models reflects a format mismatch with the standard answer-extraction rule of the GSM8K harness, while Flex.\ Acc.\ extracts the answer from \texttt{\textbackslash boxed\{\}} or the last number in the response and is thus robust to the output format.} While vanilla dLLM decoding is less than half as fast as either AR model, EB-Decode runs about $2.4\times$ faster than the faster of the two, and the remaining \ed{accuracy} gap to Qwen3-8B in Flex.\ Acc.\ stems \ed{mostly} from the base model rather than the decoding method. \mg{Beyond this matched setting, AR models benefit from recent serving systems such as vLLM~\citep{kwon2023efficient}, where Qwen3-8B and LLaMA3-8B-Instruct reach $5689.70$ and $4965.19$ TPS through paged KV-cache memory management, continuous batching, and optimized attention kernels~\citep{dao2022flashattention}, whereas serving frameworks for dLLMs such as dInfer~\citep{ma2025dinfer} are still in early development.}}

\FloatBarrier
\section{Conclusion}

Diffusion large language models promise parallel decoding but still suffer from inefficient inference under fixed block sizes and fixed confidence thresholds. To understand the source of this inefficiency, we identify two empirical phenomena: a staircase entropy pattern across token positions and an early convergence phenomenon across denoising steps. Building on these observations, we propose EB-Decode, a plug-in framework that replaces the fixed block size and the fixed confidence threshold with two lightweight learnable routers, LBS and LPS, while keeping the base dLLM frozen. Experiments on three open-source dLLMs and four benchmarks show that EB-Decode delivers up to \textbf{$18.76\times$} throughput improvement over the vanilla decoder with comparable accuracy, offering a practical step toward narrowing the efficiency gap between dLLMs and autoregressive models.

{
\bibliographystyle{plainnat}
\bibliography{ref}

\begin{thebibliography}{64}
\providecommand{\natexlab}[1]{#1}
\providecommand{\url}[1]{\texttt{#1}}
\expandafter\ifx\csname urlstyle\endcsname\relax
  \providecommand{\doi}[1]{doi: #1}\else
  \providecommand{\doi}{doi: \begingroup \urlstyle{rm}\Url}\fi

\bibitem[Arriola et~al.(2025)Arriola, Gokaslan, Chiu, Yang, Qi, Han, Sahoo, and
  Kuleshov]{arriola2025block}
Marianne Arriola, Aaron Gokaslan, Justin~T Chiu, Zhihan Yang, Zhixuan Qi, Jiaqi
  Han, Subham~Sekhar Sahoo, and Volodymyr Kuleshov.
\newblock Block diffusion: Interpolating between autoregressive and diffusion
  language models.
\newblock \emph{arXiv preprint arXiv:2503.09573}, 2025.

\bibitem[Austin et~al.(2021{\natexlab{a}})Austin, Johnson, Ho, Tarlow, and Van
  Den~Berg]{austin2021structured}
Jacob Austin, Daniel~D Johnson, Jonathan Ho, Daniel Tarlow, and Rianne Van
  Den~Berg.
\newblock Structured denoising diffusion models in discrete state-spaces.
\newblock \emph{Advances in neural information processing systems},
  34:\penalty0 17981--17993, 2021{\natexlab{a}}.

\bibitem[Austin et~al.(2021{\natexlab{b}})Austin, Odena, Nye, Bosma,
  Michalewski, Dohan, Jiang, Cai, Terry, Le, et~al.]{austin2021program}
Jacob Austin, Augustus Odena, Maxwell Nye, Maarten Bosma, Henryk Michalewski,
  David Dohan, Ellen Jiang, Carrie Cai, Michael Terry, Quoc Le, et~al.
\newblock Program synthesis with large language models.
\newblock \emph{arXiv preprint arXiv:2108.07732}, 2021{\natexlab{b}}.

\bibitem[Bao et~al.(2025)Bao, Chen, Xu, and Shang]{bao2025learning}
Wenrui Bao, Zhiben Chen, Dan Xu, and Yuzhang Shang.
\newblock Learning to parallel: Accelerating diffusion large language models
  via learnable parallel decoding.
\newblock \emph{arXiv preprint arXiv:2509.25188}, 2025.

\bibitem[Bie et~al.(2025)Bie, Cao, Chen, Du, Gong, Gong, Gu, Hu, Huang, Lan,
  Li, Li, Li, Li, Liu, Liu, Lu, Lu, Ma, Tan, Wei, Wen, Xing, Zhang, Zhao,
  Zheng, Zhou, Zhou, Zhou, Zhu, and Zhuang]{bie2025llada2}
Tiwei Bie, Maosong Cao, Kun Chen, Lun Du, Mingliang Gong, Zhuochen Gong, Yanmei
  Gu, Jiaqi Hu, Zenan Huang, Zhenzhong Lan, Chengxi Li, Chongxuan Li, Jianguo
  Li, Zehuan Li, Huabin Liu, Ling Liu, Guoshan Lu, Xiaocheng Lu, Yuxin Ma,
  Jianfeng Tan, Lanning Wei, Ji-Rong Wen, Yipeng Xing, Xiaolu Zhang, Junbo
  Zhao, Da~Zheng, Jun Zhou, Junlin Zhou, Zhanchao Zhou, Liwang Zhu, and Yihong
  Zhuang.
\newblock Llada2.0: Scaling up diffusion language models to 100b, 2025.
\newblock URL \url{https://arxiv.org/abs/2512.15745}.

\bibitem[Chen et~al.(2021{\natexlab{a}})Chen, Tworek, Jun, Yuan, Pinto, Kaplan,
  Edwards, Burda, Joseph, Brockman, et~al.]{chen2021evaluating}
Mark Chen, Jerry Tworek, Heewoo Jun, Qiming Yuan, Henrique Ponde De~Oliveira
  Pinto, Jared Kaplan, Harri Edwards, Yuri Burda, Nicholas Joseph, Greg
  Brockman, et~al.
\newblock Evaluating large language models trained on code.
\newblock \emph{arXiv preprint arXiv:2107.03374}, 2021{\natexlab{a}}.

\bibitem[Chen et~al.(2021{\natexlab{b}})Chen, Cheng, Wang, Gan, Wang, and
  Liu]{chen2021earlybert}
Xiaohan Chen, Yu~Cheng, Shuohang Wang, Zhe Gan, Zhangyang Wang, and Jingjing
  Liu.
\newblock Earlybert: Efficient bert training via early-bird lottery tickets.
\newblock In \emph{Proceedings of the 59th Annual Meeting of the Association
  for Computational Linguistics and the 11th International Joint Conference on
  Natural Language Processing (Volume 1: Long Papers)}, pages 2195--2207,
  2021{\natexlab{b}}.

\bibitem[Chen et~al.(2025{\natexlab{a}})Chen, Huang, Guo, Wei, He, Zhang, Li,
  Chen, et~al.]{chen2025dpad}
Xinhua Chen, Sitao Huang, Cong Guo, Chiyue Wei, Yintao He, Jianyi Zhang, Hai
  Li, Yiran Chen, et~al.
\newblock Dpad: Efficient diffusion language models with suffix dropout.
\newblock \emph{arXiv preprint arXiv:2508.14148}, 2025{\natexlab{a}}.

\bibitem[Chen et~al.(2025{\natexlab{b}})Chen, Fang, Ma, Yu, and
  Wang]{chen2025dparallel}
Zigeng Chen, Gongfan Fang, Xinyin Ma, Ruonan Yu, and Xinchao Wang.
\newblock dparallel: Learnable parallel decoding for dllms.
\newblock \emph{arXiv preprint arXiv:2509.26488}, 2025{\natexlab{b}}.

\bibitem[Cheng et~al.(2025)Cheng, Bian, Liu, Zhang, Yao, Tian, Wang, Guo, Chen,
  Qi, et~al.]{cheng2025sdar}
Shuang Cheng, Yihan Bian, Dawei Liu, Linfeng Zhang, Qian Yao, Zhongbo Tian,
  Wenhai Wang, Qipeng Guo, Kai Chen, Biqing Qi, et~al.
\newblock Sdar: A synergistic diffusion-autoregression paradigm for scalable
  sequence generation.
\newblock \emph{arXiv preprint arXiv:2510.06303}, 2025.

\bibitem[Cobbe et~al.(2021)Cobbe, Kosaraju, Bavarian, Chen, Jun, Kaiser,
  Plappert, Tworek, Hilton, Nakano, et~al.]{cobbe2021training}
Karl Cobbe, Vineet Kosaraju, Mohammad Bavarian, Mark Chen, Heewoo Jun, Lukasz
  Kaiser, Matthias Plappert, Jerry Tworek, Jacob Hilton, Reiichiro Nakano,
  et~al.
\newblock Training verifiers to solve math word problems.
\newblock \emph{arXiv preprint arXiv:2110.14168}, 2021.

\bibitem[Dao et~al.(2022)Dao, Fu, Ermon, Rudra, and
  R{\'e}]{dao2022flashattention}
Tri Dao, Dan Fu, Stefano Ermon, Atri Rudra, and Christopher R{\'e}.
\newblock Flashattention: Fast and memory-efficient exact attention with
  io-awareness.
\newblock \emph{Advances in neural information processing systems},
  35:\penalty0 16344--16359, 2022.

\bibitem[Deschenaux and Gulcehre(2024)]{deschenaux2024SDTT}
Justin Deschenaux and Caglar Gulcehre.
\newblock Beyond autoregression: Fast llms via self-distillation through time.
\newblock \emph{arXiv preprint arXiv:2410.21035}, 2024.

\bibitem[Frankle and Carbin(2018)]{frankle2018lottery}
Jonathan Frankle and Michael Carbin.
\newblock The lottery ticket hypothesis: Finding sparse, trainable neural
  networks.
\newblock \emph{arXiv preprint arXiv:1803.03635}, 2018.

\bibitem[{Google DeepMind}(2025)]{gemini_diffusion}
{Google DeepMind}.
\newblock Gemini diffusion.
\newblock \url{https://deepmind.google/models/gemini-diffusion/}, 2025.

\bibitem[Grattafiori et~al.(2024)Grattafiori, Dubey, Jauhri, Pandey, Kadian,
  Al-Dahle, Letman, Mathur, Schelten, Vaughan, et~al.]{grattafiori2024llama}
Aaron Grattafiori, Abhimanyu Dubey, Abhinav Jauhri, Abhinav Pandey, Abhishek
  Kadian, Ahmad Al-Dahle, Aiesha Letman, Akhil Mathur, Alan Schelten, Alex
  Vaughan, et~al.
\newblock The llama 3 herd of models.
\newblock \emph{arXiv preprint arXiv:2407.21783}, 2024.

\bibitem[Graves(2016)]{graves2016adaptive}
Alex Graves.
\newblock Adaptive computation time for recurrent neural networks.
\newblock \emph{arXiv preprint arXiv:1603.08983}, 2016.

\bibitem[Gu et~al.(2024)Gu, Fu, Liu, Shen, Lin, and Wang]{gu-etal-2024-light}
Naibin Gu, Peng Fu, Xiyu Liu, Bowen Shen, Zheng Lin, and Weiping Wang.
\newblock Light-{PEFT}: Lightening parameter-efficient fine-tuning via early
  pruning.
\newblock In Lun-Wei Ku, Andre Martins, and Vivek Srikumar, editors,
  \emph{Findings of the Association for Computational Linguistics: ACL 2024},
  pages 7528--7541, Bangkok, Thailand, August 2024. Association for
  Computational Linguistics.
\newblock \doi{10.18653/v1/2024.findings-acl.447}.
\newblock URL \url{https://aclanthology.org/2024.findings-acl.447/}.

\bibitem[Hayakawa et~al.(2024)Hayakawa, Takida, Imaizumi, Wakaki, and
  Mitsufuji]{hayakawa2024Di4C}
Satoshi Hayakawa, Yuhta Takida, Masaaki Imaizumi, Hiromi Wakaki, and Yuki
  Mitsufuji.
\newblock Distillation of discrete diffusion through dimensional correlations.
\newblock \emph{arXiv preprint arXiv:2410.08709}, 2024.

\bibitem[Hong et~al.(2025)Hong, Yu, Ye, Huang, Zheng, Zhang, Wang, and
  Yao]{hong2025wide}
Feng Hong, Geng Yu, Yushi Ye, Haicheng Huang, Huangjie Zheng, Ya~Zhang, Yanfeng
  Wang, and Jiangchao Yao.
\newblock Wide-in, narrow-out: Revokable decoding for efficient and effective
  dllms.
\newblock \emph{arXiv preprint arXiv:2507.18578}, 2025.

\bibitem[Hu et~al.(2025)Hu, Meng, Akhauri, Abdelfattah, Seo, Zhang, and
  Gupta]{hu2025flashdlm}
Zhanqiu Hu, Jian Meng, Yash Akhauri, Mohamed~S Abdelfattah, Jae-sun Seo, Zhiru
  Zhang, and Udit Gupta.
\newblock Flashdlm: Accelerating diffusion language model inference via
  efficient kv caching and guided diffusion.
\newblock \emph{arXiv preprint arXiv:2505.21467}, 2025.

\bibitem[Huang et~al.(2025)Huang, Liu, Liu, Yan, Wang, Chen, and
  Xiao]{huang2025pcsampler}
Pengcheng Huang, Shuhao Liu, Zhenghao Liu, Yukun Yan, Shuo Wang, Zulong Chen,
  and Tong Xiao.
\newblock Pc-sampler: Position-aware calibration of decoding bias in masked
  diffusion models.
\newblock \emph{arXiv preprint arXiv:2508.13021}, 2025.

\bibitem[Israel et~al.(2025)Israel, Broeck, and Grover]{israel2025accelerating}
Daniel Israel, Guy Van~den Broeck, and Aditya Grover.
\newblock Accelerating diffusion llms via adaptive parallel decoding.
\newblock \emph{arXiv preprint arXiv:2506.00413}, 2025.

\bibitem[Kang et~al.(2025)Kang, Galim, Oh, Lee, Zeng, Zhang, Hooper, Hu, Koo,
  Cho, et~al.]{kang2025parallelbench}
Wonjun Kang, Kevin Galim, Seunghyuk Oh, Minjae Lee, Yuchen Zeng, Shuibai Zhang,
  Coleman Hooper, Yuezhou Hu, Hyung~Il Koo, Nam~Ik Cho, et~al.
\newblock Parallelbench: Understanding the trade-offs of parallel decoding in
  diffusion llms.
\newblock \emph{arXiv preprint arXiv:2510.04767}, 2025.

\bibitem[Khanna et~al.(2025)Khanna, Kharbanda, Li, Varma, Wang, Birnbaum, Luo,
  Miraoui, Palrecha, Ermon, et~al.]{khanna2025mercury}
Samar Khanna, Siddhant Kharbanda, Shufan Li, Harshit Varma, Eric Wang, Sawyer
  Birnbaum, Ziyang Luo, Yanis Miraoui, Akash Palrecha, Stefano Ermon, et~al.
\newblock Mercury: Ultra-fast language models based on diffusion.
\newblock \emph{arXiv e-prints}, pages arXiv--2506, 2025.

\bibitem[Kwon et~al.(2023)Kwon, Li, Zhuang, Sheng, Zheng, Yu, Gonzalez, Zhang,
  and Stoica]{kwon2023efficient}
Woosuk Kwon, Zhuohan Li, Siyuan Zhuang, Ying Sheng, Lianmin Zheng, Cody~Hao Yu,
  Joseph Gonzalez, Hao Zhang, and Ion Stoica.
\newblock Efficient memory management for large language model serving with
  pagedattention.
\newblock In \emph{Proceedings of the 29th symposium on operating systems
  principles}, pages 611--626, 2023.

\bibitem[Li et~al.(2024)Li, Beeching, Tunstall, Lipkin, Soletskyi, Huang,
  Rasul, Yu, Jiang, Shen, Qin, Dong, Zhou, Fleureau, Lample, and
  Polu]{numina_math_datasets}
Jia Li, Edward Beeching, Lewis Tunstall, Ben Lipkin, Roman Soletskyi,
  Shengyi~Costa Huang, Kashif Rasul, Longhui Yu, Albert Jiang, Ziju Shen, Zihan
  Qin, Bin Dong, Li~Zhou, Yann Fleureau, Guillaume Lample, and Stanislas Polu.
\newblock Numinamath.
\newblock
  \url{https://github.com/project-numina/aimo-progress-prize/blob/main/report/numina_dataset.pdf},
  2024.

\bibitem[Li et~al.(2025)Li, Zhou, Muhtar, Yin, Yan, Shen, Vosoughi, and
  Liu]{li2025diffusion}
Pengxiang Li, Yefan Zhou, Dilxat Muhtar, Lu~Yin, Shilin Yan, Li~Shen, Soroush
  Vosoughi, and Shiwei Liu.
\newblock Diffusion language models know the answer before decoding.
\newblock \emph{arXiv preprint arXiv:2508.19982}, 2025.

\bibitem[Li et~al.(2026)Li, Muhtar, Chen, Yin, and Liu]{li2026diffusion}
Pengxiang Li, Dilxat Muhtar, Tianlong Chen, Lu~Yin, and Shiwei Liu.
\newblock Why diffusion language models struggle with truly parallel
  (non-autoregressive) decoding?
\newblock \emph{arXiv preprint arXiv:2602.23225}, 2026.

\bibitem[Liang et~al.(2026)Liang, Wang, Chen, Sun, Wu, Yu, Liu, Barsoum, Liu,
  and Jha]{liang2026cd4lm}
Yihao Liang, Ze~Wang, Hao Chen, Ximeng Sun, Jialian Wu, Xiaodong Yu, Jiang Liu,
  Emad Barsoum, Zicheng Liu, and Niraj~K Jha.
\newblock Cd4lm: Consistency distillation and adaptive decoding for diffusion
  language models.
\newblock \emph{arXiv preprint arXiv:2601.02236}, 2026.

\bibitem[Lightman et~al.(2023)Lightman, Kosaraju, Burda, Edwards, Baker, Lee,
  Leike, Schulman, Sutskever, and Cobbe]{lightman2023lets}
Hunter Lightman, Vineet Kosaraju, Yura Burda, Harri Edwards, Bowen Baker, Teddy
  Lee, Jan Leike, John Schulman, Ilya Sutskever, and Karl Cobbe.
\newblock Let's verify step by step.
\newblock \emph{arXiv preprint arXiv:2305.20050}, 2023.

\bibitem[Ling et~al.(2017)Ling, Yogatama, Dyer, and Blunsom]{ling2017program}
Wang Ling, Dani Yogatama, Chris Dyer, and Phil Blunsom.
\newblock Program induction by rationale generation: Learning to solve and
  explain algebraic word problems.
\newblock In \emph{Proceedings of the 55th annual meeting of the association
  for computational linguistics (volume 1: Long papers)}, pages 158--167, 2017.

\bibitem[Liu et~al.(2025)Liu, Yang, Zhang, Chen, Zou, Wei, Wang, and
  Zhang]{liu2025dllm}
Zhiyuan Liu, Yicun Yang, Yaojie Zhang, Junjie Chen, Chang Zou, Qingyuan Wei,
  Shaobo Wang, and Linfeng Zhang.
\newblock dllm-cache: Accelerating diffusion large language models with
  adaptive caching.
\newblock \emph{arXiv preprint arXiv:2506.06295}, 2025.

\bibitem[Lou et~al.(2023)Lou, Meng, and Ermon]{lou2023discrete}
Aaron Lou, Chenlin Meng, and Stefano Ermon.
\newblock Discrete diffusion modeling by estimating the ratios of the data
  distribution.
\newblock \emph{arXiv preprint arXiv:2310.16834}, 2023.

\bibitem[Lu et~al.(2025)Lu, Chen, Karashima, Wang, Fujiki, and
  Fan]{lu2025adablock}
Guanxi Lu, Hao~Mark Chen, Yuto Karashima, Zhican Wang, Daichi Fujiki, and
  Hongxiang Fan.
\newblock Adablock-dllm: Semantic-aware diffusion llm inference via adaptive
  block size.
\newblock \emph{arXiv preprint arXiv:2509.26432}, 2025.

\bibitem[Luo et~al.(2026)Luo, Li, Wen, and Zhang]{luo2026dsb}
Lizhuo Luo, Shenggui Li, Yonggang Wen, and Tianwei Zhang.
\newblock Dsb: Dynamic sliding block scheduling for diffusion llms.
\newblock \emph{arXiv preprint arXiv:2602.05992}, 2026.

\bibitem[Luxembourg et~al.(2025)Luxembourg, Permuter, and
  Nachmani]{luxembourg2025plan}
Omer Luxembourg, Haim Permuter, and Eliya Nachmani.
\newblock Plan for speed--dilated scheduling for masked diffusion language
  models.
\newblock \emph{arXiv preprint arXiv:2506.19037}, 2025.

\bibitem[Ma et~al.(2025{\natexlab{a}})Ma, Yu, Fang, and Wang]{ma2025dkv}
Xinyin Ma, Runpeng Yu, Gongfan Fang, and Xinchao Wang.
\newblock dkv-cache: The cache for diffusion language models.
\newblock \emph{arXiv preprint arXiv:2505.15781}, 2025{\natexlab{a}}.

\bibitem[Ma et~al.(2025{\natexlab{b}})Ma, Du, Wei, Chen, Xu, Wang, Feng, Lu,
  Liu, Qi, Zhang, Tao, Feng, Jiang, Xu, Huang, Zhuang, Xu, Hu, Lan, Zhao, Li,
  and Zheng]{ma2025dinfer}
Yuxin Ma, Lun Du, Lanning Wei, Kun Chen, Qian Xu, Kangyu Wang, Guofeng Feng,
  Guoshan Lu, Lin Liu, Xiaojing Qi, Xinyuan Zhang, Zhen Tao, Haibo Feng, Ziyun
  Jiang, Ying Xu, Zenan Huang, Yihong Zhuang, Haokai Xu, Jiaqi Hu, Zhenzhong
  Lan, Junbo Zhao, Jianguo Li, and Da~Zheng.
\newblock dinfer: An efficient inference framework for diffusion language
  models.
\newblock \emph{arXiv preprint arXiv:2510.08666}, 2025{\natexlab{b}}.

\bibitem[Nie et~al.(2025)Nie, Zhu, You, Zhang, Ou, Hu, Zhou, Lin, Wen, and
  Li]{nie2025large}
Shen Nie, Fengqi Zhu, Zebin You, Xiaolu Zhang, Jingyang Ou, Jun Hu, Jun Zhou,
  Yankai Lin, Ji-Rong Wen, and Chongxuan Li.
\newblock Large language diffusion models.
\newblock \emph{arXiv preprint arXiv:2502.09992}, 2025.

\bibitem[Peng et~al.(2025)Peng, Liu, Dong, Cheng, Li, Tang, Wang, and
  Zhao]{peng2025efficient}
Han Peng, Peiyu Liu, Zican Dong, Daixuan Cheng, Junyi Li, Yiru Tang, Shuo Wang,
  and Wayne~Xin Zhao.
\newblock How efficient are diffusion language models? a critical examination
  of efficiency evaluation practices.
\newblock \emph{arXiv preprint arXiv:2510.18480}, 2025.

\bibitem[Sahoo et~al.(2024)Sahoo, Arriola, Schiff, Gokaslan, Marroquin, Chiu,
  Rush, and Kuleshov]{sahoo2024simple}
Subham~S Sahoo, Marianne Arriola, Yair Schiff, Aaron Gokaslan, Edgar Marroquin,
  Justin~T Chiu, Alexander Rush, and Volodymyr Kuleshov.
\newblock Simple and effective masked diffusion language models.
\newblock \emph{Advances in Neural Information Processing Systems},
  37:\penalty0 130136--130184, 2024.

\bibitem[Schuster et~al.(2022)Schuster, Fisch, Gupta, Dehghani, Bahri, Tran,
  Tay, and Metzler]{schuster2022confident}
Tal Schuster, Adam Fisch, Jai Gupta, Mostafa Dehghani, Dara Bahri, Vinh Tran,
  Yi~Tay, and Donald Metzler.
\newblock Confident adaptive language modeling.
\newblock \emph{Advances in Neural Information Processing Systems},
  35:\penalty0 17456--17472, 2022.

\bibitem[Shi et~al.(2024)Shi, Han, Wang, Doucet, and
  Titsias]{shi2024simplified}
Jiaxin Shi, Kehang Han, Zhe Wang, Arnaud Doucet, and Michalis Titsias.
\newblock Simplified and generalized masked diffusion for discrete data.
\newblock \emph{Advances in neural information processing systems},
  37:\penalty0 103131--103167, 2024.

\bibitem[Song et~al.(2025)Song, Zhang, Luo, Gao, Xia, Luo, Li, Yang, Yu, Qu,
  et~al.]{song2025seed}
Yuxuan Song, Zheng Zhang, Cheng Luo, Pengyang Gao, Fan Xia, Hao Luo, Zheng Li,
  Yuehang Yang, Hongli Yu, Xingwei Qu, et~al.
\newblock Seed diffusion: A large-scale diffusion language model with
  high-speed inference.
\newblock \emph{arXiv preprint arXiv:2508.02193}, 2025.

\bibitem[Teerapittayanon et~al.(2016)Teerapittayanon, McDanel, and
  Kung]{teerapittayanon2016branchynet}
Surat Teerapittayanon, Bradley McDanel, and Hsiang-Tsung Kung.
\newblock Branchynet: Fast inference via early exiting from deep neural
  networks.
\newblock In \emph{2016 23rd international conference on pattern recognition
  (ICPR)}, pages 2464--2469. IEEE, 2016.

\bibitem[Wang et~al.(2025{\natexlab{a}})Wang, Schiff, Sahoo, and
  Kuleshov]{wang2025remasking}
Guanghan Wang, Yair Schiff, Subham~Sekhar Sahoo, and Volodymyr Kuleshov.
\newblock Remasking discrete diffusion models with inference-time scaling.
\newblock \emph{arXiv preprint arXiv:2503.00307}, 2025{\natexlab{a}}.

\bibitem[Wang et~al.(2025{\natexlab{b}})Wang, Fang, Jing, Shen, Shen, Wang,
  Ouyang, Chen, and Shen]{wang2025time}
Wen Wang, Bozhen Fang, Chenchen Jing, Yongliang Shen, Yangyi Shen, Qiuyu Wang,
  Hao Ouyang, Hao Chen, and Chunhua Shen.
\newblock Time is a feature: Exploiting temporal dynamics in diffusion language
  models.
\newblock \emph{arXiv preprint arXiv:2508.09138}, 2025{\natexlab{b}}.

\bibitem[Wang et~al.(2025{\natexlab{c}})Wang, Xu, Jin, Jin, Zhang, and
  Deng]{wang2025diffusion}
Xu~Wang, Chenkai Xu, Yijie Jin, Jiachun Jin, Hao Zhang, and Zhijie Deng.
\newblock Diffusion llms can do faster-than-ar inference via discrete diffusion
  forcing.
\newblock \emph{arXiv preprint arXiv:2508.09192}, 2025{\natexlab{c}}.

\bibitem[Wei et~al.(2022)Wei, Wang, Schuurmans, Bosma, Xia, Chi, Le, Zhou,
  et~al.]{wei2022chain}
Jason Wei, Xuezhi Wang, Dale Schuurmans, Maarten Bosma, Fei Xia, Ed~Chi, Quoc~V
  Le, Denny Zhou, et~al.
\newblock Chain-of-thought prompting elicits reasoning in large language
  models.
\newblock \emph{Advances in neural information processing systems},
  35:\penalty0 24824--24837, 2022.

\bibitem[Wei et~al.(2025)Wei, Zhang, Liu, Liu, and Zhang]{wei2025accelerating}
Qingyan Wei, Yaojie Zhang, Zhiyuan Liu, Dongrui Liu, and Linfeng Zhang.
\newblock Accelerating diffusion large language models with slowfast sampling:
  The three golden principles.
\newblock \emph{arXiv preprint arXiv:2506.10848}, 2025.

\bibitem[Whalen et~al.(2025)Whalen, Du, You, Li, Li, and Lin]{whalen2025early}
Lexington Whalen, Zhenbang Du, Haoran You, Chaojian Li, Sixu Li, and
  Yingyan~Celine Lin.
\newblock Early-bird diffusion: Investigating and leveraging timestep-aware
  early-bird tickets in diffusion models for efficient training.
\newblock \emph{arXiv preprint arXiv:2504.09606}, 2025.

\bibitem[Wu et~al.(2025)Wu, Zhang, Xue, Liu, Diao, Zhu, Luo, Han, and
  Xie]{wu2025fast}
Chengyue Wu, Hao Zhang, Shuchen Xue, Zhijian Liu, Shizhe Diao, Ligeng Zhu, Ping
  Luo, Song Han, and Enze Xie.
\newblock Fast-dllm: Training-free acceleration of diffusion llm by enabling kv
  cache and parallel decoding.
\newblock \emph{arXiv preprint arXiv:2505.22618}, 2025.

\bibitem[Xin et~al.(2020)Xin, Tang, Lee, Yu, and Lin]{xin2020deebert}
Ji~Xin, Raphael Tang, Jaejun Lee, Yaoliang Yu, and Jimmy Lin.
\newblock Deebert: Dynamic early exiting for accelerating bert inference.
\newblock In \emph{Proceedings of the 58th annual meeting of the association
  for computational linguistics}, pages 2246--2251, 2020.

\bibitem[Yang et~al.(2025)Yang, Li, Yang, Zhang, Hui, Zheng, Yu, Gao, Huang,
  Lv, Zheng, Liu, Zhou, Huang, Hu, Ge, Wei, Lin, Tang, Yang, Tu, Zhang, Yang,
  Yang, Zhou, Zhou, Lin, Dang, Bao, Yang, Yu, Deng, Li, Xue, Li, Zhang, Wang,
  Zhu, Men, Gao, Liu, Luo, Li, Tang, Yin, Ren, Wang, Zhang, Ren, Fan, Su,
  Zhang, Zhang, Wan, Liu, Wang, Cui, Zhang, Zhou, and Qiu]{qwen3}
An~Yang, Anfeng Li, Baosong Yang, Beichen Zhang, Binyuan Hui, Bo~Zheng, Bowen
  Yu, Chang Gao, Chengen Huang, Chenxu Lv, Chujie Zheng, Dayiheng Liu, Fan
  Zhou, Fei Huang, Feng Hu, Hao Ge, Haoran Wei, Huan Lin, Jialong Tang, Jian
  Yang, Jianhong Tu, Jianwei Zhang, Jianxin Yang, Jiaxi Yang, Jing Zhou,
  Jingren Zhou, Junyang Lin, Kai Dang, Keqin Bao, Kexin Yang, Le~Yu, Lianghao
  Deng, Mei Li, Mingfeng Xue, Mingze Li, Pei Zhang, Peng Wang, Qin Zhu, Rui
  Men, Ruize Gao, Shixuan Liu, Shuang Luo, Tianhao Li, Tianyi Tang, Wenbiao
  Yin, Xingzhang Ren, Xinyu Wang, Xinyu Zhang, Xuancheng Ren, Yang Fan, Yang
  Su, Yichang Zhang, Yinger Zhang, Yu~Wan, Yuqiong Liu, Zekun Wang, Zeyu Cui,
  Zhenru Zhang, Zhipeng Zhou, and Zihan Qiu.
\newblock Qwen3 technical report.
\newblock \emph{arXiv preprint arXiv:2505.09388}, 2025.

\bibitem[Ye et~al.(2025)Ye, Xie, Zheng, Gao, Wu, Jiang, Li, and
  Kong]{ye2025dream}
Jiacheng Ye, Zhihui Xie, Lin Zheng, Jiahui Gao, Zirui Wu, Xin Jiang, Zhenguo
  Li, and Lingpeng Kong.
\newblock Dream 7b: Diffusion large language models.
\newblock \emph{arXiv preprint arXiv:2508.15487}, 2025.

\bibitem[You et~al.(2019)You, Li, Xu, Fu, Wang, Chen, Baraniuk, Wang, and
  Lin]{you2019drawing}
Haoran You, Chaojian Li, Pengfei Xu, Yonggan Fu, Yue Wang, Xiaohan Chen,
  Richard~G Baraniuk, Zhangyang Wang, and Yingyan~Celine Lin.
\newblock Drawing early-bird tickets: Towards more efficient training of deep
  networks.
\newblock \emph{arXiv preprint arXiv:1909.11957}, 2019.

\bibitem[You et~al.(2022)You, Lu, Zhou, Fu, and Lin]{you2022early}
Haoran You, Zhihan Lu, Zijian Zhou, Yonggan Fu, and Yingyan Lin.
\newblock Early-bird gcns: Graph-network co-optimization towards more efficient
  gcn training and inference via drawing early-bird lottery tickets.
\newblock In \emph{Proceedings of the AAAI Conference on Artificial
  Intelligence}, volume~36, pages 8910--8918, 2022.

\bibitem[Zhang et~al.(2026{\natexlab{a}})Zhang, Zhang, Han, Shi, He, Li, Wang,
  Xu, Srivastava, Pavlovic, et~al.]{zhang2026t3d}
Tunyu Zhang, Xinxi Zhang, Ligong Han, Haizhou Shi, Xiaoxiao He, Zhuowei Li, Hao
  Wang, Kai Xu, Akash Srivastava, Vladimir Pavlovic, et~al.
\newblock T3d: Few-step diffusion language models via trajectory
  self-distillation with direct discriminative optimization.
\newblock \emph{arXiv preprint arXiv:2602.12262}, 2026{\natexlab{a}}.

\bibitem[Zhang et~al.(2026{\natexlab{b}})Zhang, Li, Zhou, Ma, Wan, Shi, Miao,
  Zhang, and Cao]{zhang2026swordsman}
Yu~Zhang, Xinchen Li, Jialei Zhou, Hongnan Ma, Zhongwei Wan, Yiwei Shi, Duoqian
  Miao, Qi~Zhang, and Longbing Cao.
\newblock Swordsman: Entropy-driven adaptive block partition for efficient
  diffusion language models.
\newblock \emph{arXiv preprint arXiv:2602.04399}, 2026{\natexlab{b}}.

\bibitem[Zheng et~al.(2025)Zheng, Chen, Mao, Liu, Zhu, and
  Zhang]{zheng2025masked}
Kaiwen Zheng, Yongxin Chen, Hanzi Mao, Ming-Yu Liu, Jun Zhu, and Qinsheng
  Zhang.
\newblock Masked diffusion models are secretly time-agnostic masked models and
  exploit inaccurate categorical sampling.
\newblock In \emph{International Conference on Learning Representations}, 2025.

\bibitem[Zhou et~al.(2020)Zhou, Xu, Ge, McAuley, Xu, and Wei]{zhou2020bert}
Wangchunshu Zhou, Canwen Xu, Tao Ge, Julian McAuley, Ke~Xu, and Furu Wei.
\newblock Bert loses patience: Fast and robust inference with early exit.
\newblock \emph{Advances in Neural Information Processing Systems},
  33:\penalty0 18330--18341, 2020.

\bibitem[Zhu et~al.(2025{\natexlab{a}})Zhu, Wang, Nie, Zhang, Wu, Hu, Zhou,
  Chen, Lin, Wen, et~al.]{zhu2025llada}
Fengqi Zhu, Rongzhen Wang, Shen Nie, Xiaolu Zhang, Chunwei Wu, Jun Hu, Jun
  Zhou, Jianfei Chen, Yankai Lin, Ji-Rong Wen, et~al.
\newblock Llada 1.5: Variance-reduced preference optimization for large
  language diffusion models.
\newblock \emph{arXiv preprint arXiv:2505.19223}, 2025{\natexlab{a}}.

\bibitem[Zhu et~al.(2025{\natexlab{b}})Zhu, You, Xing, Huang, Liu, Zhuang, Lu,
  Wang, Wang, Wei, et~al.]{zhu2025lladamoe}
Fengqi Zhu, Zebin You, Yipeng Xing, Zenan Huang, Lin Liu, Yihong Zhuang,
  Guoshan Lu, Kangyu Wang, Xudong Wang, Lanning Wei, et~al.
\newblock Llada-moe: A sparse moe diffusion language model.
\newblock \emph{arXiv preprint arXiv:2509.24389}, 2025{\natexlab{b}}.

\end{thebibliography}
}

\clearpage
\appendix

\section{Detailed dLLM Backgrounds}
\label{app:dllm_related}

This section extends the dLLM literature summary in \ed{Sec.~\ref{sec:related}} of the main text with additional context on representative methods.

D3PM~\citep{austin2021structured} generalizes discrete diffusion using structured transition matrices over the token vocabulary. Subsequent works refine the masked diffusion paradigm. MDLM~\citep{sahoo2024simple} derives a Rao-Blackwellized objective that reduces to a weighted mixture of masked language modeling losses. MD4~\citep{shi2024simplified} further simplifies and generalizes the continuous-time objective. SEDD~\citep{lou2023discrete} proposes a score-entropy objective. \citet{zheng2025masked} shows that masked diffusion training and sampling are time-agnostic. Together, these advances bring masked dLLMs to likelihood performance comparable to AR models.
LLaDA~\citep{nie2025large} performs on par with LLaMA3 at the billion-parameter scale. Subsequent extensions add variance-reduced preference optimization~\citep{zhu2025llada}, sparse experts~\citep{zhu2025lladamoe}, and AR-to-diffusion conversion that scales to 100B parameters~\citep{bie2025llada2}. Dream-7B~\citep{ye2025dream} introduces AR-based LLM initialization and context-adaptive, token-level noise rescheduling.
Semi-AR methods such as BD3-LMs~\citep{arriola2025block} and SDAR~\citep{cheng2025sdar} combine the coherence of AR generation with the parallelism of diffusion. They maintain AR dependencies across blocks while sampling tokens in parallel within each block.

\section{Block-wise Decoding for dLLMs}
\label{app:dllm_prelim}

This section details the block-wise decoding recipe summarized in Sec.~\ref{sec:preliminaries of dLLMs}, reusing the discrete-diffusion notation introduced therein, including the vocabulary $\mathcal{V}$, the mask symbol $\texttt{[MASK]}$, the mask predictor $p_\theta$, the clean response sequence $\mathbf{x}_0$ with its position-$i$ token $x^i_0$, and the partially denoised sequence $\mathbf{x}_t$ at noise level $t \in [0,1]$. We further write $x^i_t$ for the token at position $i$ in $\mathbf{x}_t$, which is either a vocabulary item in $\mathcal{V}$ or the mask symbol $\texttt{[MASK]}$.

\textbf{Inference-time denoising trajectory.}
At inference time, given a prompt $c$, decoding starts from a fully masked response sequence and iteratively applies the reverse process from $t = 1$ to $t = 0$ over a sequence of denoising steps. The prompt $c$ is concatenated with the response and processed jointly by the bidirectional Transformer~\citep{nie2025large}, so that the predictive distribution at every masked position conditions on both $\mathbf{x}_t$ and $c$. The final output is the fully unmasked sequence $\mathbf{x}_0$.

\textbf{Block partition and in-block prediction.}
The response is divided into contiguous blocks $\mathcal{B}_1, \mathcal{B}_2, \ldots$ of fixed size $B$, indexed by $j$, and processed from left to right. Within the active block $\mathcal{B}_j$, the model predicts all masked positions in parallel, commits a subset of tokens, and remasks the remaining uncertain ones for further refinement. The per-position top-1 confidence at masked position $i \in \mathcal{B}_j$ over candidate vocabulary item $v \in \mathcal{V}$ follows the same definition as in the main text:
\begin{equation}
    \kappa^i = \max_{v \in \mathcal{V}} p_\theta\bigl(x^i_0 = v \,\bigm|\, \mathbf{x}_t,\, c\bigr).
    \label{eq:app-confidence}
\end{equation}
Given a confidence threshold $\tau$ or an optional per-step commit budget $m$, the standard block update rule is
\begin{equation}
    x^i_t \leftarrow
    \begin{cases}
        \displaystyle\argmax_{v \in \mathcal{V}} p_\theta\bigl(x^i_0 = v \,\bigm|\, \mathbf{x}_t,\, c\bigr),
        & \kappa^i \ge \tau \;\; \text{or} \;\; i \in \mathrm{Top}\text{-}m(\mathcal{B}_j), \\
        \texttt{[MASK]}, & \text{otherwise,}
    \end{cases}
    \label{eq:prelim-update}
\end{equation}
where $\mathrm{Top}\text{-}m(\mathcal{B}_j)$ denotes the $m$ masked positions in $\mathcal{B}_j$ with the highest top-1 confidence at the current step. Once all positions in $\mathcal{B}_j$ have been committed, the active block advances to $\mathcal{B}_{j+1}$ and the same procedure repeats. Throughout this recipe, the block size $B$, the threshold $\tau$, and the budget $m$ remain fixed regardless of how token difficulty varies along the sequence, which is precisely the limitation that motivates our EB-Decode framework. Building on the same block-wise procedure and keeping the base dLLM frozen, EB-Decode replaces the two static design choices with learnable counterparts. Specifically, LBS replaces the fixed block size $B$ with a learnable block selector as described in Sec.~\ref{sec:adablock}, while LPS replaces the static commit rule with a position-aware, learnable acceptance strategy as described in Sec.~\ref{sec:LPS}.

\section{Detailed Setup for Learnable Block Size}
\label{app:lbs_setup}

\subsection{Router Architecture}
\label{app:router_arch}

The LBS router $\mathcal{R}_\phi$ is a small Transformer encoder with 600k parameters (not including the frozen token embedding table $W_e$) that maps per-masked-position features to scalar inclusion logits.

\textbf{Entropy branch.}
The normalized entropy sequence for all masked tokens \ed{$\tilde{H} = \{\tilde{H}^i : i \in \mathcal{M}\}$} is first zero-padded to the generation length $\pp{L}$ and then processed by a two-layer MLP $\pp{F_{\mathrm{ent}}}$:
$\text{Linear}(1 \to 64) \to \text{GELU} \to \text{Linear}(64 \to 64) \to \text{LayerNorm}$,
producing a 64-dimensional entropy embedding.

\textbf{Token branch.}
The top-1 predicted token IDs \ed{$\hat{x} = \{\hat{x}^i : i \in \mathcal{M}\}$ are} looked up in the frozen base-model token embedding table $W_e$, zero-padded to $\pp{L}$, and projected to $\mathbb{R}^{64}$ via a learned linear layer $W_{\text{tok}} \in \mathbb{R}^{64 \times \pp{d_{\mathrm{base}}}}$.

\textbf{Fusion and positional embedding.}
The two 64-dimensional embeddings are concatenated to form a 128-dimensional vector, then passed through a fusion layer $W_{\text{combine}} \in \mathbb{R}^{128 \times 128}$.
The final input embedding to the Transformer encoder is calculated as:
\begin{equation}
    \pp{Z_\phi}
    =
    W_{\text{combine}}
    \!\left[\,
    \pp{F_{\mathrm{ent}}}\!\left(\tilde{H}\right)
    \;;\;
    W_{\text{tok}}\,W_e\bigl(\hat{x}\bigr)
    \,\right]
    +
    \pp{E_{\mathrm{pos}}}, \pp{Z_\phi} \in \mathbb{R}^{\pp{L} \times 128}
    \label{eq:lbs_feat}
\end{equation}
where $\pp{E_{\mathrm{pos}}} \in \mathbb{R}^{\pp{L} \times 128}$ is a learnable positional embedding.

\textbf{Transformer encoder.}
The sequence $\pp{Z_\phi}$ is then fed into a 2-layer Transformer encoder with hidden dimension \ed{$d_\phi{=}128$}, 4 attention heads, FFN width 256, and GELU activations.

\textbf{Output head.}
A linear head $\mathbb{R}^{128} \to \mathbb{R}^1$ maps the encoder output to logits $O_\phi \in \mathbb{R}^{\pp{L} \times 1}$.
The padding is removed before returning, yielding an output of shape $(|\mathcal{M}|, 1)$. 
The final inclusion probability at position $i$ is \pp{$p^i_\phi = \sigma(O^i_\phi)$}.

\subsection{Training Procedure}
\label{app:lbs_train_proc}

The full per-step training procedure for the LBS router, summarized in the main text and the simplified \ed{Alg.~\ref{alg:eb_train}}, is given \ed{in Alg.~\ref{alg:lbs_train}}. \new{Training is performed on $8\times$A100 GPUs.}

\algnewcommand{\Notation}{\item[\textbf{Notation:}]}

\begin{algorithm}[H]
\caption{LBS Router Training}
\label{alg:lbs_train}
\begin{algorithmic}[1]
\Require dLLM-generated response dataset $\mathcal{D}$, frozen dLLM $\mathcal{F}_\theta$, router $\mathcal{R}_\phi$, generation length $L$, block length $L_B$, mask ratio $r \in (0,1]$, learning rate $\eta_{\mathrm{lr}}$
\Notation
\Statex \hspace{1em} $\tilde{H}(p) = (-\sum_{v \in \mathcal{V}} p(v)\log p(v))/\log |\mathcal{V}|$
\Statex \hspace{1em} $\hat{x}(p) = \arg\max_{x} p(x)$
\Statex \hspace{1em} $\sigma(p) = 1 / (1+e^{-p})$
\For{$x \in \mathcal{D}$}
    \State Sample $n \sim \{0,\, 1 , \ldots,\, L / L_B - 1\}$
    \State $x_{\mathrm{masked}} \leftarrow \text{Blockwise-Masking}(x, r, n, L, L_B)$; $\mathcal{M} \leftarrow \{i : x_{\mathrm{masked}}^i = \texttt{[MASK]}\}$
    \For{each $i \in \mathcal{M}$}
    \State $p^i \gets \mathrm{softmax}(\mathcal{F}_\theta(x_{\mathrm{masked}})^i)$
    \State $\ell^i_{\mathrm{CE}} = \text{CrossEntropy}(p^i, x^i)$ \Comment{per-token CE against ground truth}
    \State $O^i_\phi \gets \mathcal{R}_\phi(\tilde{H}(p^i), \hat{x}(p^i))$
    \EndFor
    \State 
    $\mathcal{L}_{\mathrm{wCE}} 
    = \frac{1}{|\mathcal{M}|}
    \displaystyle\sum_{i \in \mathcal{M}}
    \sigma(O^i_\phi) \cdot \ell^i_{\mathrm{CE}}$
    \State $\mathcal{L}_{\mathrm{Reg}} = 
    \frac{1}{|\mathcal{M}|}\,\displaystyle\sum_{i \in \mathcal{M}} \sigma(O^i_\phi)$
    \State $\mathcal{L}_{\mathrm{LBS}} \gets \mathcal{L}_{\mathrm{wCE}} - \mathcal{L}_{\mathrm{Reg}}$
    \State $\mathcal{R}_\phi \leftarrow \mathcal{R}_\phi - \eta_{\mathrm{lr}}\,\nabla_\phi\,\mathcal{L}_{\mathrm{LBS}}$ \Comment{update only router; $\mathcal{F}_\theta$ is frozen}
\EndFor
\State \Return $\mathcal{R}_\phi$
\end{algorithmic}
\end{algorithm}

\new{\textbf{Training stability.}}
\begin{revblock}
Tab.~\ref{tab:train_stability} reports the training and validation losses of the LBS router. The training loss is negative because Eq.~\eqref{eq:lbs_loss} subtracts the regularization term from the weighted cross-entropy loss. Both losses drop sharply in the first epoch, after which the changes are much smaller, indicating stable training.
\end{revblock}

\begin{table}[ht]
\centering
\caption{Training and validation losses of the LBS router across epochs.}
\label{tab:train_stability}
\setlength{\tabcolsep}{4pt}
\small
\begin{tabular}{c|cc}
\toprule
\textbf{Epoch} & \textbf{Train loss} & \textbf{Val loss} \\
\midrule
0 & 0.0569  & 0.0764  \\
1 & $-$0.337 & $-$0.291 \\
3 & $-$0.364 & $-$0.378 \\
6 & $-$0.359 & $-$0.396 \\
\bottomrule
\end{tabular}
\end{table}

\section{Detailed Setup for Learnable Parallel Sampling}
\label{app:lps_setup}

\subsection{Router Architecture}
\label{app:lps_arch}

The LPS router $\mathcal{R}_\psi$ is a small Transformer encoder that operates over all positions of the active block jointly. Taking the per-position input $(f^i, \mathrm{pos}^i)$ defined in Sec.~\ref{sec:LPS}, the input embedding \pp{$Z_\psi$} is a linear projection followed by LayerNorm and GELU, mapping each feature vector into a hidden space of dimension \ed{$d_\psi=32$}\pp{, so that $Z_\psi \in \mathbb{R}^{|\mathcal{B}| \times 32}$}. The encoder then processes \pp{$Z_\psi$} with $2$ pre-norm Transformer blocks with $4$ attention heads, a feed-forward width of $64$, GELU activations, and dropout of $0.1$. A final linear head projects each hidden state to a scalar logit $O^i_\psi$\pp{, from which the commit probability is $p^i_\psi = \sigma(O^i_\psi)$}, giving a total of roughly $27$k trainable parameters. This keeps the router cost negligible compared with the base dLLM forward pass.

\subsection{Trace Generation}
\label{app:lps_trace}

To remove any train/test distribution skew, we collect traces under the exact decoding configuration that the router will face at inference. For each experiment we run the corresponding frozen base dLLM with generation length $512$, base block size $32$, and confidence threshold $\tau{=}0.9$, matching the inference setting in Sec.~\ref{sec:exp_setup}. Since the distribution differs across base models, we train a separate router for each base model.
Every record captures, at each position $i$ of the active block, the four feature components $f^i_t = (\kappa^i_t, \tilde{H}^i_t, \Delta^i_t, \rho_t)$, the predicted token $\hat{x}^i_t$, the correctness label $y^i_t = \mathbf{1}[\hat{x}^i_t = x^i_0]$ obtained by comparing against the completed sequence $x_0$, and the mask indicator $m^i_t$ which shows which position is still masked. The replay pass follows the oracle progress rule described in Sec.~\ref{sec:LPS}, so each intermediate state stays consistent with $x_0$ and the labels $y^i_t$ provide a clean supervision signal without external annotation.

\begin{table}[t]
\centering
\caption{Training and validation losses and validation recall of the LPS router across epochs.}
\label{tab:train_stability_lps}
\setlength{\tabcolsep}{4pt}
\small
\begin{tabular}{c|ccc}
\toprule
\textbf{Epoch} & \textbf{Train loss} & \textbf{Val loss} & \textbf{Val recall} \\
\midrule
0    & 0.6225 & 0.3987 & 0.6471 \\
20   & 0.3699 & 0.3640 & 0.6998 \\
60   & 0.3677 & 0.3622 & 0.7042 \\
119  & 0.3669 & 0.3618 & \textbf{0.7052} \\
\bottomrule
\end{tabular}
\end{table}

\begin{algorithm}[htb]
\caption{\ed{LPS Router Training}}
\label{alg:lps-training}
\begin{algorithmic}[1]

\Require  Frozen dLLM $\mathcal{F}_\theta$; prompt corpus $\mathcal{D}$;
        LBS Router $\mathcal{R}_\phi$; LPS Router $\mathcal{R}_\psi$;
         negative weight $w^- = 2$; target False-Positive rate $\eta\!\in\!(0,1)$;
         learning rate $\eta_{\mathrm{lr}}$; early-stop patience \ed{$P_{\mathrm{patience}}$}.
\Ensure  Router parameters $\mathcal{R}_\psi$ and acceptance threshold $\tau_\psi$.

\Statex \textbf{Stage 1: build the trace dataset $\mathcal{T}$.}

\State $\mathcal{T} \gets \emptyset$
       \Comment{$\mathcal{T}$ stores per-position records $(f^i_t,\,\mathrm{pos}^i,\,y^i_t)$.}
\ForAll{prompt $q \in \mathcal{D}$}
\Comment{Run LBS decoding on $q$ to obtain $x_0$ and the per-step trajectory.}
  \ForAll{step $t$ and masked position $i \in \mathcal{B}$ at step $t$}
    \State $p^i_t \gets \mathrm{softmax}\bigl(\mathcal{F}_\theta(\mathbf{x}_t)\bigr)_i$
           \Comment{predictive distribution at position $i$ from $\mathbf{x}_t$.}
    \State $\kappa^i_t \gets \max_v p^i_t(v)$,\;
           $\hat{x}^i_t \gets \arg\max_v p^i_t(v)$
           \Comment{top-1 confidence and predicted token.}
    \State $\tilde{H}^i_t,\, \Delta^i_t,\, \rho_t \gets$ entropy, top-1/top-2 gap, block-wise mask ratio
    \State $f^i_t \gets (\kappa^i_t,\, \tilde{H}^i_t,\, \Delta^i_t,\, \rho_t)$,\;
           $\mathrm{pos}^i \gets \ed{\sfrac{(i - b)}{|\mathcal{B}|}}$
    \State $y^i_t \gets \mathbf{1}[\hat{x}^i_t = x^i_0]$
           \Comment{correctness label vs.\ $x_0$.}
    \State \textbf{Append} $(f^i_t,\,\mathrm{pos}^i,\,y^i_t)$ to $\mathcal{T}$
  \EndFor
\EndFor

\Statex \textbf{Stage 2: LPS router $\mathcal{R}_\psi$ training.}
\Repeat
  \ForAll{mini-batch $\mathcal{S} \subset \mathcal{T}_{\mathrm{train}}$}
    \State $O^i_\psi \gets \mathcal{R}_\psi(f^i_t,\, \mathrm{pos}^i_t)$ for each record $(f^i_t,\,\mathrm{pos}^i,\,y^i_t) \in \mathcal{S}$
\State $\pp{\mathcal{L}_{\mathrm{LPS}}} \gets \dfrac{1}{|\mathcal{S}|}\displaystyle\sum_{\pp{(i,t)} \in \mathcal{S}} \pp{w^i_t}\,\mathrm{BCE}(\sigma(O^i_\psi),\,y^i_t)$,\quad
where $\pp{w^i_t} = \begin{cases} 1, & y^i_t = 1 \\ w^-, & y^i_t = 0 \end{cases}$
          \Comment{\pp{mini-batch form of} Eq.~\ed{\eqref{eq:lps_loss}}}
    \State $\mathcal{R}_\psi \gets \mathcal{R}_\psi - \eta_{\mathrm{lr}}\,\nabla \pp{\mathcal{L}_{\mathrm{LPS}}}$
  \EndFor
  \State Evaluate validation recall on $\mathcal{T}_{\mathrm{val}}$ under $\mathrm{FP} \leq \eta$
\Until{validation recall has not improved for \ed{$P_{\mathrm{patience}}$} consecutive epochs}
\State get $\mathcal{T}_{\mathrm{val}}
        \quad\text{s.t.}\quad
        |\mathrm{FP}_{\mathcal{T}_{\mathrm{val}}}(\tau_\psi) - \eta| \le \delta$
        \Comment{within $\eta\pm 0.5$ pp false-positive rate on validation set.}
\State \Return $(\mathcal{R}_\psi,\,\tau_\psi)$
\end{algorithmic}
\end{algorithm}

\subsection{Training Procedure}
\label{app:lps_training}

Valid trajectories are split at the trace level into $80\%$ training, $10\%$ validation, and $10\%$ test sets, with the split deterministic under seed $42$ so that every router variant sees identical data. The positions outside the active block are masked out of both the loss and the attention computation to prevent boundary artifacts from contaminating the supervision.

Optimization uses AdamW with an initial learning rate of $1\mathrm{e}{-3}$, weight decay $0.01$, gradient clipping at norm $1.0$, and automatic mixed precision. The schedule consists of a linear warmup over $0.5\%$ of the total steps followed by a per-epoch geometric decay with ratio $0.97$. We train with a batch size of $512$ step records for up to $2000$ epochs, with early stopping triggered when the validation recall at the false-positive budget fails to improve for $100$ consecutive \ed{epochs}.

The training objective is the masked weighted binary cross-entropy defined in Eq.~\ed{\eqref{eq:lps_loss}}, where negative samples carry a weight of $w^{-}=2$. We use asymmetric weights because a false positive commits a wrong token irreversibly, while a false negative only delays acceptance to a later step. Premature acceptance therefore has a higher cost.

We tune the LPS threshold $\tau_\psi$ on the validation set at the end of each improving epoch by searching over a $250$-point grid in $[0.8,\,0.9999]$ and keeping the largest recall under a false-positive-rate budget of $\eta=10^{-2}$. The full procedure is summarized in Alg.~\ref{alg:lps-training}.

\new{\textbf{Training stability.}}
\begin{revblock}
As shown in Tab.~\ref{tab:train_stability_lps}, the training and validation losses converge smoothly and the validation recall rises steadily. \pp{The validation loss stays close to the training loss throughout, indicating no overfitting.}
\end{revblock}

\section{EB-Decode Inference Procedure}
\label{app:eb_infer_proc}

The complete inference procedure that combines LBS with \ed{LPS is given in Alg.~\ref{alg:lbs_lps_infer}}. The simplified high-level view is in Alg.~\ref{alg:eb_infer}.

\begin{algorithm}[H]
\caption{Inference with LBS Block Selection and LPS Parallel Commit}
\label{alg:lbs_lps_infer}
\begin{algorithmic}[1]
\Require Prompt $p$, frozen dLLM $\mathcal{F}_\theta$; LBS router $\mathcal{R}_\phi$; LPS router $\mathcal{R}_\psi$; generation length $L$; min block size $L_{\min}$; fallback block size $L_B$; max gap $g_{\max}$; confidence threshold $\tau$; LPS acceptance threshold $\tau_\psi$

\State $x \leftarrow [p \;|\; \underbrace{\texttt{[MASK]}, \ldots, \texttt{[MASK]}}_{L}]$;\;\; $\mathcal{B} \leftarrow \emptyset$
\While{$\exists\, i$ such that $x^i = \texttt{[MASK]}$}
    \State $p_t \leftarrow \mathrm{softmax}\bigl(\mathcal{F}_\theta(x)\bigr)$;\;\; $\mathcal{M} \leftarrow \{i : x^i = \texttt{[MASK]}\}$ \Comment{get distribution and masked places}
    \For{each $i \in \mathcal{M}$}
        \State $\kappa^i \leftarrow \max p^i_t$;\;\; $\hat{x}^i \leftarrow \arg\max p^i_t$
        \State Compute $\tilde{H}^i$, $\Delta^i$, $\rho$;\;\; 
        \Comment{compute entropy, top-1/top-2 gap, block-wise mask ratio}
        \State $f^i \leftarrow (\kappa^i,\, \tilde{H}^i,\, \Delta^i,\, \rho)$ \Comment{LPS feature tuple}
    \EndFor
    \If{$\mathcal{B} = \emptyset$} \Comment{LBS: select a new active block}
        \If{$|\mathcal{M}| \leq L_{\min}$}
            \State $\mathcal{B} \leftarrow$ $\mathcal{M}$ \Comment{size fallback}
        \Else
            \State $\mathcal{C} \leftarrow \{i \in \mathcal{M} : \sigma(\mathcal{R}_\phi(\tilde{H}^i,\, \hat{x}^i)) > 0.5\ \wedge \ \hat{x}^i \neq \text{EOS} \}$ \Comment{router + EOS removal}
            \State $\mathcal{B} \leftarrow \mathrm{MaxGapFilter}(\mathcal{C},\, g_{\max})$ \Comment{spatial coherence}
            \If{$|\mathcal{B}| < L_{\min}$}
                \State $\mathcal{B} \leftarrow$ first $L_B$ positions of $\mathcal{M}$ \Comment{min-tokens fallback}
            \EndIf
        \EndIf
    \EndIf
    \State $O^i_\psi \leftarrow \mathcal{R}_\psi(f^i,\,  \ed{\sfrac{(i - b)}{|\mathcal{B}|}})$ for $i \in \mathcal{B}$  \Comment{LPS router, $b$ is the left boundary index of $\mathcal{B}$}
    \State $\mathcal{A} \leftarrow \bigl\{i \in \mathcal{B} : x^i = \texttt{[MASK]} \,\wedge\, \bigl(\kappa^i \geq \tau \,\vee\, \pp{p^i_\psi} > \tau_\psi\bigr)\bigr\}$ \Comment{union accept rule}
    \If{$\mathcal{A} = \emptyset$}
        \State $\mathcal{A} \leftarrow \{\arg\max_{i \in \mathcal{B},\, x^i = \texttt{[MASK]}} \kappa^i\}$ \Comment{anti-stall fallback}
    \EndIf
    \State $x^i \leftarrow \hat{x}^i$ for \ed{$i \in \mathcal{A}$} \Comment{parallel commit}
    \If{$x^i \neq \texttt{[MASK]}$ for all $i \in \mathcal{B}$}
        \State $\mathcal{B} \leftarrow \emptyset$ \Comment{If $\mathcal{B}$ resolved, move to Line 9}
    \EndIf
\EndWhile
\State \Return $x$
\end{algorithmic}
\end{algorithm}

\section{\texorpdfstring{\ed{Experimental Setup}}{Experimental Setup}}
\label{app:exp_setting}
\pp{\textbf{Decoding configuration.}}
All methods share the same inference configuration of generation length $512$, base block size $32$, and base confidence threshold $\tau{=}0.9$.
\new{No method in Tab.~\ref{tab:main_results} uses a KV cache. In the official implementation of Fast-dLLM~\citep{wu2025fast}, the KV cache is recomputed with one full-sequence forward pass at the start of each block and reused within that block, whereas the vanilla LLaDA decoding loop maintains no cache.}

\pp{\textbf{EB-Decode hyperparameters.}}
For LBS in our EB-Decode, we set $g_{\max} = 5$ for all \ed{three} dLLMs, and $L_{\min} = 8, 8, 16$ for LLaDA-8B-Instruct, Dream-v0-Instruct-7B, LLaDA-1.5 respectively. When not paired with LPS, LBS adopts confidence-based decoding following Fast-dLLM~\citep{wu2025fast}.
For LPS, we choose the acceptance threshold $\tau_\psi$ under a false-positive-rate budget of $1\% \pm 0.5$ \% points on the validation set, giving \ed{$\tau_\psi = 0.89$, $0.90$, and $0.87$} for LLaDA-8B-Instruct, Dream-v0-Instruct-7B, LLaDA-1.5 respectively.

\pp{\textbf{Throughput measurement.}}
\new{All methods, including every baseline, are measured with the HuggingFace \texttt{transformers} backend at batch size $1$, following prior dLLM acceleration works~\citep{lu2025adablock,bao2025learning,chen2025dparallel}, and throughput (TPS) is measured per GPU as the number of generated tokens divided by the end-to-end latency.}

\pp{\textbf{Hardware.}}
All runs are performed on $4 \times$ NVIDIA H100 GPUs unless specified otherwise.
\new{The experiments in \ed{Tabs.~\ref{tab:abl_cache} and~\ref{tab:transfer}} are run on $4\times$ NVIDIA A100 GPUs, and those in Tab.~\ref{tab:blockforward} on NVIDIA A100 GPUs with a maximum generation length of $512$.}

\section{\texorpdfstring{\new{Positioning Against Prior Adaptive dLLM Decoding Methods}}{Positioning Against Prior Adaptive dLLM Decoding Methods}}
\label{app:positioning}

\begin{revblock}
To make explicit how EB-Decode differs from prior adaptive dLLM decoding methods, Tab.~\ref{tab:positioning} compares them along two axes: \emph{where to decode}, i.e., how the block is formed, and \emph{when to commit}, i.e., how tokens are accepted. Prior methods either fix the block size or form contiguous blocks with hand-designed heuristics, and most of them commit tokens with a static confidence threshold. EB-Decode is the only one that learns both axes and supports non-contiguous, variable-length blocks, so hard tokens can be deferred and resolved with richer right-side context once the surrounding easy tokens are decoded (Fig.~\ref{fig:teaser}).
\end{revblock}

\begin{table}[H]
\centering
\caption{Positioning against prior dLLM decoding methods along the two decoding axes.}
\label{tab:positioning}
\setlength{\tabcolsep}{4pt}
\small
\begin{tabular}{lccc}
\toprule
\textbf{Method} & \textbf{Where to decode} & \textbf{When to commit} & \tabincell{c}{\textbf{Non-contiguous}\\\textbf{blocks}} \\
\midrule
Fast-dLLM~\citep{wu2025fast}          & fixed size          & confidence threshold           & no  \\
AdaBlock-dLLM~\citep{lu2025adablock}  & delimiter heuristic & confidence threshold           & no  \\
Swordsman~\citep{zhang2026swordsman}  & entropy heuristic   & confidence threshold           & no  \\
Learn2PD~\citep{bao2025learning}      & fixed size          & learned (MLP, fixed length)    & no  \\
\rowcolor[RGB]{235,243,252} EB-Decode (ours) & \textbf{learned} & \textbf{learned (Transformer, variable length)} & \textbf{yes} \\
\bottomrule
\end{tabular}
\end{table}

\section{\texorpdfstring{\new{Entropy at the Sequence Tail}}{Entropy at the Sequence Tail}}
\label{app:eoszone}

\begin{revblock}
Fig.~\ref{fig:entropy}~(a) shows a persistent band of low entropy at high token positions across all decoding steps, which may appear at odds with the overall rising-entropy trend. This region corresponds to the EOS zone at the tail of the sequence. \ed{We report in Tab.~\ref{tab:eoszone} the three most frequent predictions at the last five positions across all decoding steps}, where \texttt{<|endoftext|>} is overwhelmingly dominant, appearing in the top-5 predictions at $96.30\%$ of the positions sampled. These positions therefore remain highly certain at almost every step, forming the persistent low-entropy band.
\end{revblock}

\begin{table}[H]
\centering
\caption{Top-3 predicted tokens at the last $5$ positions of the sequence, aggregated over all decoding steps on LLaDA-8B-Instruct. ``Top-5 occurrence'' is the fraction of sampled positions at which the token appears among the top-5 predictions.}
\label{tab:eoszone}
\setlength{\tabcolsep}{4pt}
\small
\begin{tabular}{c l cc}
\toprule
\textbf{Rank} & \textbf{Token} & \textbf{Avg. prob.\,(\%)} & \textbf{Top-5 occurrence\,(\%)} \\
\midrule
1 & \texttt{<|endoftext|>} & 28.00 & 96.30 \\
2 & \texttt{<|eot\_id|>}   & 18.00 & 45.70 \\
3 & \texttt{.}             & 2.70  & 24.70 \\
\bottomrule
\end{tabular}
\end{table}

\section{Limitations}
\label{app:limitations}

We outline the main limitations of EB-Decode and the directions that we leave for future work.

\textbf{Generation Length.} Our main results are reported at generation lengths of 256 and 512 tokens, which match the standard configurations used in prior dLLM acceleration works~\citep{wu2025fast,lu2025adablock,bao2025learning}. As shown in Tab.~\ref{tab:abl_genlen}, EB-Decode already exhibits a scaling trend in which the speedup grows from $L{=}256$ to $L{=}512$, suggesting that the framework remains effective as the sequence becomes longer. A more comprehensive study at substantially longer generation lengths is a natural extension we plan to pursue in future work.

\new{\textbf{Serving Conditions.} Our throughput is measured at batch size $1$, following prior dLLM acceleration works~\citep{wu2025fast,lu2025adablock,bao2025learning}. Evaluation under realistic serving conditions with continuous batching and system-level optimizations is left to future work, as it depends as much on dLLM serving infrastructure as on the decoding algorithm.}

\new{\textbf{Method Complexity.} EB-Decode introduces two routers and several hyperparameters, namely $g_{\max}$, $L_{\min}$, $\tau$, and $\tau_\psi$, which makes it more involved to deploy and reuse as a baseline than a single-threshold heuristic. Each component contributes to the final performance (Tabs.~\ref{tab:abl_lbs}--\ref{tab:selector_control}), and the trained routers are largely reusable without retraining (Sec.~\ref{sec:generality}), but reducing the number of moving parts remains a worthwhile direction.}

\end{document}